\documentclass{article}

\usepackage{PRIMEarxiv}

\usepackage[utf8]{inputenc} % allow utf-8 input
\usepackage[T1]{fontenc}    % use 8-bit T1 fonts
\usepackage[acronym, nonumberlist]{glossaries}
\usepackage{svg}
\usepackage{enumitem}
\usepackage{multicol}
\usepackage{hyperref}       % hyperlinks
\usepackage{url}            % simple URL typesetting
\usepackage{booktabs}       % professional-quality tables
\usepackage{amsfonts}       % blackboard math symbols
\usepackage{nicefrac}       % compact symbols for 1/2, etc.
\usepackage{microtype}      % microtypography
\usepackage{lipsum}
\usepackage{fancyhdr}       % header
\usepackage{graphicx}       % graphics
\graphicspath{{media/}}     % organize your images and other figures under media/ folder
\usepackage{esint}

\newacronym{SDSU}{SDSU}{San Diego State University}
\newacronym{CPL}{CPL}{Computational Physics Lab}
\newacronym{US}{US}{United States}
\newacronym{STEM}{STEM}{Science, Technology, Engineering and Mathematics}
\newacronym{TUd}{TU/d}{Delft University of Technology}
\newacronym{CAP}{CAP}{Climate Action Plan}
\newacronym{LEED}{LEED}{Leadership in Energy and Environmental Design}
\newacronym{CSU}{CSU}{California State University}
\newacronym{AM}{AM}{Additive Manufacturing}
\newacronym{ML}{ML}{Machine Learning}
\newacronym{CFD}{CFD}{Computational Fluid Dynamics}
\newacronym{RGB}{RGB}{Red, Blue and Green}
\newacronym{HSV}{HSV}{Hue, Saturation and Brightness}
\newacronym{DFT}{DFT}{Discrete Fourier Transform}
\newacronym{IDFT}{IDFT}{Inverse Discrete Fourier Transform}
\newacronym{FS}{FS}{Fourier Series}
\newacronym{FT}{FT}{Fourier Transform}
\newacronym{IFT}{IFT}{Inverse Fourier Transform}
\newacronym{SLM}{SLM}{Selective Laser Melting}
\newacronym{LPBF}{LPBF}{Laser Power Bed Fusion}
\newacronym{MSc}{MSc}{Master of Science}
\newacronym{RMS}{RMS}{Root Mean Square}

\newglossary[slg]{symbolslist}{syi}{syj}{Symbols} % create add. symbolslist
\newglossaryentry{sym:f}{
    type={symbolslist},
    name=\ensuremath{f},
    description={Field in physical space}}
\newglossaryentry{sym:ftilde}{
    type={symbolslist},
    name=\ensuremath{\Tilde{f}},
    description={Scaled field in physical space}}
\newglossaryentry{sym:fhat}{
    type={symbolslist},
    name=\ensuremath{\hat{f}},
    description={Field in spectral space}}
\newglossaryentry{sym:x}{
    type={symbolslist},
    name=\ensuremath{x},
    description={Coordinate on the x-axis}} 
\newglossaryentry{sym:y}{
    type={symbolslist},
    name=\ensuremath{y},
    description={Coordinate on the y-axis}} 
\newglossaryentry{sym:z}{
    type={symbolslist},
    name=\ensuremath{z},
    description={Coordinate on the z-axis}} 
\newglossaryentry{sym:xtilde}{
    type={symbolslist},
    name=\ensuremath{\Tilde{x}},
    description={Scaled coordinate on the x-axis}} 
\newglossaryentry{sym:ytilde}{
    type={symbolslist},
    name=\ensuremath{\Tilde{y}},
    description={Scaled coordinate on the y-axis}} 
\newglossaryentry{sym:L}{
    type={symbolslist},
    name=\ensuremath{L},
    description={Domain length in physical space}} 
\newglossaryentry{sym:Ltilde}{
    type={symbolslist},
    name=\ensuremath{\Tilde{L}},
    description={Scaled domain length in physical space}} 
\newglossaryentry{sym:R}{
    type={symbolslist},
    name=\ensuremath{R},
    description={Red part of the pixel \gls{RGB} representation}} 
\newglossaryentry{sym:Rprime}{
    type={symbolslist},
    name=\ensuremath{R'},
    description={Normalized red part of the pixel \gls{RGB} representation}} 
\newglossaryentry{sym:G}{
    type={symbolslist},
    name=\ensuremath{G},
    description={Green part of the pixel \gls{RGB} representation}} 
\newglossaryentry{sym:Gprime}{
    type={symbolslist},
    name=\ensuremath{G'},
    description={Normalized green part of the pixel \gls{RGB} representation}} 
\newglossaryentry{sym:B}{
    type={symbolslist},
    name=\ensuremath{B},
    description={Blue part of the pixel \gls{RGB} representation}} 
\newglossaryentry{sym:Bprime}{
    type={symbolslist},
    name=\ensuremath{B'},
    description={Normalized blue part of the pixel \gls{RGB} representation}} 
\newglossaryentry{sym:I}{
    type={symbolslist},
    name=\ensuremath{I},
    description={Pixel resolution in the x-direction}} 
\newglossaryentry{sym:K}{
    type={symbolslist},
    name=\ensuremath{K},
    description={Pixel resolution in the y-direction}} 
\newglossaryentry{sym:H}{
    type={symbolslist},
    name=\ensuremath{H},
    description={Hue value of a pixel}} 
\newglossaryentry{sym:S}{
    type={symbolslist},
    name=\ensuremath{S},
    description={Saturation value of a pixel}} 
\newglossaryentry{sym:V}{
    type={symbolslist},
    name=\ensuremath{V},
    description={Brightness value of a pixel}} 
\newglossaryentry{sym:Hprime}{
    type={symbolslist},
    name=\ensuremath{H'},
    description={Normalized hue value of a pixel}}
\newglossaryentry{sym:N}{
    type={symbolslist},
    name=\ensuremath{N},
    description={Amount of samples in the x-direction}}
\newglossaryentry{sym:M}{
    type={symbolslist},
    name=\ensuremath{M},
    description={Amount of samples in the y-direction}}
\newglossaryentry{sym:error}{
    type={symbolslist},
    name=\ensuremath{\varepsilon},
    description={Error}}
\newglossaryentry{sym:k}{
    type={symbolslist},
    name=\ensuremath{k},
    description={Wave number}}
\newglossaryentry{sym:kvector}{
    type={symbolslist},
    name=\ensuremath{\mathbf{k}},
    description={Wave number vector}}
\newglossaryentry{sym:kvectormagnitude}{
    type={symbolslist},
    name=\ensuremath{|\mathbf{k}|},
    description={Wave number vector magnitude}}
\newglossaryentry{sym:kvectortilde}{
    type={symbolslist},
    name=\ensuremath{\Tilde{\mathbf{k}}},
    description={Scaled wave number vector}}
\newglossaryentry{sym:imag}{
    type={symbolslist},
    name=\ensuremath{\underline{i}},
    description={Imaginary number: $\underline{i} = \sqrt{-1}$}}
\newglossaryentry{sym:s}{
    type={symbolslist},
    name=\ensuremath{s},
    description={Scaling factor of the x-axis}}
\newglossaryentry{sym:r}{
    type={symbolslist},
    name=\ensuremath{r},
    description={Scaling factor of the y-axis}}
\newglossaryentry{sym:Real}{
    type={symbolslist},
    name=\ensuremath{\mathbb{R}},
    description={Real numbers}}
\newglossaryentry{sym:Integers}{
    type={symbolslist},
    name=\ensuremath{\mathbb{Z}},
    description={Integers}}
\newglossaryentry{sym:Complex}{
    type={symbolslist},
    name=\ensuremath{\mathbb{C}},
    description={Complex numbers}}
\newglossaryentry{sym:kmagnitudetilde}{
    type={symbolslist},
    name=\ensuremath{|\mathbf{\Tilde{k}}|},
    description={Scaled wave vector magnitude}}
\newglossaryentry{sym:E}{
    type={symbolslist},
    name=\ensuremath{E},
    description={Energy Spectrum}}
\newglossaryentry{sym:sigma}{
    type={symbolslist},
    name=\ensuremath{\sigma},
    description={Integration constant}}
\newglossaryentry{sym:theta}{
    type={symbolslist},
    name=\ensuremath{\theta},
    description={Random numbers: $\in [-\pi, \pi)$}}
\newglossaryentry{sym:alpha}{
    type={symbolslist},
    name=\ensuremath{\mathcal{\alpha}},
    description={Random Number Generator function}}
\newglossaryentry{sym:Phi}{
    type={symbolslist},
    name=\ensuremath{\Phi},
    description={Random numbers: $\in [0, 2 \pi)$}}
\newglossaryentry{sym:evector}{
    type={symbolslist},
    name=\ensuremath{\mathbf{e}},
    description={Spectral unit vector basis}}
\newglossaryentry{sym:bvector}{
    type={symbolslist},
    name=\ensuremath{\mathbf{b}},
    description={Computational unit vector basis}}
\newglossaryentry{sym:T}{
    type={symbolslist},
    name=\ensuremath{\mathbf{T}},
    description={Transformation matrix}}
\newglossaryentry{sym:identity}{
    type={symbolslist},
    name=\ensuremath{\mathbf{I}},
    description={Identity matrix}}
\newglossaryentry{sym:a}{
    type={symbolslist},
    name=\ensuremath{a},
    description={Fourier Series coefficients of the sine terms in the real form}}
\newglossaryentry{sym:b}{
    type={symbolslist},
    name=\ensuremath{b},
    description={Fourier Series coefficients of the cosine terms in the real form}}
\newglossaryentry{sym:c}{
    type={symbolslist},
    name=\ensuremath{c},
    description={Fourier Series coefficients in the complex form}}
\newglossaryentry{sym:n}{
    type={symbolslist},
    name=\ensuremath{n},
    description={Wave mode number}}
\newglossaryentry{sym:m}{
    type={symbolslist},
    name=\ensuremath{m},
    description={Wave mode number}}
\newglossaryentry{sym:Rtensor}{
    type={symbolslist},
    name=\ensuremath{\mathbf{R}},
    description={Two-point correlation tensor in physical space}}
\newglossaryentry{sym:Rtensorhat}{
    type={symbolslist},
    name=\ensuremath{\hat{\mathbf{R}}},
    description={Two-point correlation tensor in spectral space}}
\newglossaryentry{sym:omega0}{
    type={symbolslist},
    name=\ensuremath{\omega_0},
    description={Fundamental frequency: $\omega_0 = \frac{2 \pi }{L}$}}
\newglossaryentry{sym:u}{
    type={symbolslist},
    name=\ensuremath{u},
    description={Velocity field in physical space}}
\newglossaryentry{sym:uhat}{
    type={symbolslist},
    name=\ensuremath{\hat{u}},
    description={Velocity field in spectral space}}
\newglossaryentry{sym:fprime}{
    type={symbolslist},
    name=\ensuremath{f'},
    description={Normalized Two-Point longitudinal correlation function}}
\newglossaryentry{sym:Re}{
    type={symbolslist},
    name=\ensuremath{Re},
    description={Reynolds number}}
\newglossaryentry{sym:rvector}{
    type={symbolslist},
    name=\ensuremath{\mathbf{r}},
    description={Separation vector in physical space}}
\newglossaryentry{sym:rvectormagnitude}{
    type={symbolslist},
    name=\ensuremath{|\mathbf{r}|},
    description={Separation vector bins in physical space}}
\newglossaryentry{sym:xvector}{
    type={symbolslist},
    name=\ensuremath{\mathbf{x}},
    description={Point vector in physical space}}
\newglossaryentry{sym:t}{
    type={symbolslist},
    name=\ensuremath{t},
    description={Time}}
\newglossaryentry{sym:uprime}{
    type={symbolslist},
    name=\ensuremath{u'},
    description={Velocity field fluctuations}}
\newglossaryentry{sym:gamma}{
    type={symbolslist},
    name=\ensuremath{\gamma},
    description={Polar coordinate}}
\newglossaryentry{sym:P}{
    type={symbolslist},
    name=\ensuremath{P},
    description={Amount of Gaussian distributions}}
    
\newglossary[slg]{superscripts}{syn}{sym}{Superscripts}
\newglossaryentry{sup:E}{
    type={superscripts},
    name=\ensuremath{E},
    description={Extracted}}
\newglossaryentry{sup:R}{
    type={superscripts},
    name=\ensuremath{R},
    description={Rogallo}} 
\newglossaryentry{sup:O}{
    type={superscripts},
    name=\ensuremath{O},
    description={Original}}
\newglossaryentry{sup:RGB}{
    type={superscripts},
    name=\ensuremath{RGB},
    description={Red, Blue, Green, pixel representation}} 
\newglossaryentry{sup:H}{
    type={superscripts},
    name=\ensuremath{H},
    description={Hue}} 
\newglossaryentry{sup:Hprime}{
    type={superscripts},
    name=\ensuremath{H'},
    description={Normalized hue}}
\newglossaryentry{sup:I}{
    type={superscripts},
    name=\ensuremath{I},
    description={Interpolated}}
\newglossaryentry{sup:F}{
    type={superscripts},
    name=\ensuremath{F},
    description={Fourier}}
\newglossaryentry{sup:V}{
    type={superscripts},
    name=\ensuremath{V},
    description={Vorticity}}
\newglossaryentry{sup:inverse}{
    type={superscripts},
    name=\ensuremath{-1},
    description={Inverse}}
\newglossaryentry{sup:transpose}{
    type={superscripts},
    name=\ensuremath{T},
    description={Transpose}}
\newglossaryentry{sup:*}{
    type={superscripts},
    name=\ensuremath{*},
    description={Complex conjugate}}
\newglossaryentry{sup:tot}{
    type={superscripts},
    name=\ensuremath{tot},
    description={Total}}
\newglossaryentry{sup:V2}{
    type={superscripts},
    name=\ensuremath{V^2},
    description={Dissipation}}
\newglossaryentry{sup:prime}{
    type={superscripts},
    name=\ensuremath{'},
    description={Normalized}}
\newglossaryentry{sup:G}{
    type={superscripts},
    name=\ensuremath{G},
    description={Gaussian}}
    
\newglossary[slg]{subscripts}{syl}{syh}{Subscripts}
\newglossaryentry{sub:i}{
    type={subscripts},
    name=\ensuremath{i},
    description={Sampling counter index in the \gls{sym:x}-axis in physical space}}
\newglossaryentry{sub:j}{
    type={subscripts},
    name=\ensuremath{j},
    description={Sampling counter index in the \gls{sym:y}-axis in physical space}}
\newglossaryentry{sub:x}{
    type={subscripts},
    name=\ensuremath{x},
    description={Along \gls{sym:x}-axis}}
\newglossaryentry{sub:y}{
    type={subscripts},
    name=\ensuremath{y},
    description={Along \gls{sym:y}-axis}}
\newglossaryentry{sub:n}{
    type={subscripts},
    name=\ensuremath{n},
    description={Counter index along the \gls{sym:x}-axis in spectral space}}
\newglossaryentry{sub:m}{
    type={subscripts},
    name=\ensuremath{m},
    description={Counter index along the \gls{sym:y}-axis in spectral space}}
\newglossaryentry{sub:max}{
    type={subscripts},
    name=\ensuremath{max},
    description={Maximum}}
\newglossaryentry{sub:min}{
    type={subscripts},
    name=\ensuremath{min},
    description={Minimum}}
\newglossaryentry{sub:l}{
    type={subscripts},
    name=\ensuremath{l},
    description={Pixel location index along the x-axis}}
\newglossaryentry{sub:k}{
    type={subscripts},
    name=\ensuremath{k},
    description={Pixel location index along the y-axis}}
\newglossaryentry{sub:plot}{
    type={subscripts},
    name=\ensuremath{plot},
    description={Plot resolution}}
\newglossaryentry{sub:be}{
    type={subscripts},
    name=\ensuremath{be},
    description={From the spectral to computational unit vector basis}}
\newglossaryentry{sub:ij}{
    type={subscripts},
    name=\ensuremath{ij},
    description={Tensor index notation}}
\newglossaryentry{sub:ii}{
    type={subscripts},
    name=\ensuremath{ii},
    description={Tensor trace index notation}}
\newglossaryentry{sub:iindex}{
    type={subscripts},
    name=\ensuremath{i},
    description={Index notation in \gls{sym:x}}}
\newglossaryentry{sub:jindex}{
    type={subscripts},
    name=\ensuremath{j},
    description={Index notation in \gls{sym:y}}}
\newglossaryentry{sub:cr}{
    type={subscripts},
    name=\ensuremath{cr},
    description={Critical}}

\makenoidxglossaries

\title{A  Synthetic Modal  Generation of   Additive Manufacturing Roughness Surfaces  from Images 
}

\author{
  T.B. Keesom, P.P. Popov, P. Dhyani, G.B. Jacobs. \\
  San Diego State University \\
  San Diego\\
}

\begin{document}
\maketitle

\begin{abstract}
%\lipsum[1]
A method to infer and synthetically extrapolate roughness fields from electron microscope scans of additively manufactured surfaces using an adaptation of Rogallo's synthetic turbulence method [R. S. Rogallo, NASA Technical Memorandum 81315, 1981] based on Fourier modes is presented. The resulting synthetic roughness fields are smooth and are compatible with grid generators in computational fluid dynamics or other numerical simulations. Unlike machine learning methods, which can require over twenty scans of surface roughness for training, the Fourier mode based method can extrapolate homogeneous synthetic roughness fields using a single physical roughness scan to any desired size and range. Five types of synthetic roughness fields are generated using an electron microscope roughness image from literature. A comparison of their spectral energy and two-point correlation spectra show that the synthetic fields closely approximate the roughness structures and spectral energy of the scan. 
\end{abstract}

% keywords can be removed

%Glossaries
%\glsaddall
%\printnoidxglossaries
%\printnoidxglossary[type=\acronymtype]%, style=long4col]

\section{Introduction}
\label{sec:Introduction}

Metal 3D printing techniques such as \gls{LPBF} are becoming more widely used in the science and engineering. A particular class of \gls{LPBF}, \gls{SLM} uses a powder bed of evenly spread metallic powder over a working area, which is fused into the desired shape by locally delivering energy to melt the powder particles together. Once melting of one layer is done, a new evenly spread layer of metallic powder is added to the work area to continue the laser meting process until the final 3D shape is completed \cite{Frazier2014, BouabbouVaudreuil2022}. 
%\autoref{fig:SLMschematic} shows a schematic of the \gls{SLM} metal \gls{3D} printing technique.
%\begin{figure}[h!]
%    \centering
%    \includegraphics[width=0.5\textwidth]{Images/SLMschematic}
%    \caption{A generic schematic of an \gls{AM} powder bed system \cite{Frazier2014}}
%    \label{fig:SLMschematic}
%\end{figure}

A well known disadvantage of \gls{AM} is its rough surface quality. \gls{SLM} part surfaces are 4 to 5 times rougher compared to machined surfaces \cite{MowerLong2016}. The \gls{SLM} roughness topography is determined by the various physical processes lying at the heart of the manufacturing process \cite{BouabbouVaudreuil2022}, an example being the steep cooling rate of the molten powder particles after the delivery of the energy by the laser \cite{AboulkhairEtAl2019}. The topography depends on variables such as laser input energy, scan speed, scan width, exposure time \cite{AdeleKniepkamp2015, PatelEtAl2020} and the printed object's orientation with respect to the laser \cite{FoxEtAl2016, KasperovichEtAl2021}. While some reduction of roughness height is possible\cite{AdeleKniepkamp2015}, eliminating the roughness structures appears difficult even with additional post-processing techniques \cite{FaveroEtAl2022}. 

Roughness topographies interact with and affect the aerodynamics of the flow over rough surfaces \cite{Chen1972, AntoniaLuxton1971, AntoniaLuxton1972}, especially if the roughness topographies extend beyond a critical roughness height \cite{BrezginEtAl2017}. The aerodynamic mixing process and heat transfer performance are affected by the interaction of the wall roughness with the flow \cite{GramespacherEtAl2021, KadivarEtAl2022}. Comparing numerical predictions with experimental measurements indicates that not all wall roughness effects on the flow are numerically captured \cite{McClainEtAl2021}. The wall roughness may affect the flow to the point that \gls{CFD} simulations lose their predictive capability and no longer yield reliable results \cite{FaveroEtAl2021}.

Modeling of the effect of roughness in CFD is complicated by the range of scales present in \gls{AM}. The large roughness structures can be represented by means of a discrete number of triangle \cite{ZhangEtAl2010, NoorianEtAl2014} square \cite{ZhangEtAl2010}, semicircle \cite{ZhangEtAl2010}, cylinder \cite{NoorianEtAl2014}, cantor curve \cite{ChenEtAl2010}, sinusoidal curve \cite{DharaiyaKandlikar2013}, conical \cite{GroceEtAl2007} or cuboid \cite{HuEtAl2003, RawoolEtAl2006, HeckPapavassiliou2013} shaped elements. The non-differentiable nature of some functions can pose challenges for higher approximations of governing equations in CFD. The block spectral meshing methods of Kapsis et al. \cite{Kapsis2019, KapsisEtAl2019} have been proposed with the aim of alleviating these issues and lowering computational cost by avoiding to resolve similar geometry areas with similar flow properties and define fixed regions of set environmental conditions. A drawback of these methods is that insufficient element-to-element resolution can yield erroneous predictions. Another method uses a collection of random fractal or Gaussian functions, generating a larger range of roughness than models utilizing only discrete roughness elements \cite{ChenEtAl2009, GuoEtAl2015}. A problem of fractal surfaces are sharp corners in the roughness topography which result in sudden local pressure drops in the flow over it \cite{ZhangEtAl2012}. Gaussian distribution surfaces do not feature such corners, but still do not capture the range of scales present in the roughness \cite{XiongChung2010, GuoEtAl2015}. Sen et al. \cite{SenEtAl2015} report that models utilizing Gaussian basis functions show a larger sum mean square error and converge more slowly than other existing models such as Fourier series and dynamic Kriging. %Unfortunately, dynamic Kriging cannot be easily adapted for the generation of new random roughness surfaces; this leaves Fourier series, which are adapted in this work from an approximation to a new surface generation method.

%Transition to ML
%\gls{ML} is also being used for the modelling of wall roughness topographies. \gls{ML} is an approach that is increasingly employed to learn behavior of complex systems. The relation between the input scan data and the output predicted roughness is highly non-linear, making the problem very suitable for \gls{ML} \cite{BouabbouVaudreuil2022}. \gls{ML} methods have therefore been used extensively in the field of \gls{SLM}. 
 
\gls{ML} approaches have shown promise in the modeling of \gls{AM} roughness geometries.
Khorasini \textit{et al.} show that artificial neural networks model surface roughness geometries of a \gls{SLM} created specimen with about half the root mean square error compared to the non-\gls{ML} Poisson and Taguchi method \cite{KhorasaniEtAl2018}.
Fotovvati \textit{et al.} and La Fé-Perdomo \textit{et al.} find similar results in a different setting, achieving root mean square errors in the order of a few percent when approximating experimental roughness geometry data using \gls{ML} methods \cite{FotovvatiEtAl2022, LaFé-PerdomoEtAl2022}.  
Most of the considered \gls{ML} models by La Fé-Perdomo \textit{et al.} predict input data roughness geometries within the experimental measurement 95\% confidence intervals \cite{LaFé-PerdomoEtAl2022}.

A disadvantage of \gls{ML} methods that they require a large amount of data for training. In \cite{KhorasaniEtAl2018, FotovvatiEtAl2022, LaFé-PerdomoEtAl2022}, between twenty-one to ninety-four experimental roughness scans are used to train the \gls{ML} models. 
%"To train and test the Neural Network, 21 and 3 experiments are used respectively", p.3770 [Khorasani et a 2018].
%
%"A total of 110 data sets were randomly split into 85% training and 15% testing data", p.126 [Fotovati 2022]
%"A total of 81 experiments (27 experiments and three replicates) are used to train/test the ML models", p. 674 [La Fe-Perdomo]
The extraction of this amount of data from images relies on access to expensive equipment such as electron microscopes and might not be practical.
Another disadvantage of \gls{ML} models is that they are uninformative beyond providing the requested output. An \gls{ML} model may yield a good approximation of roughness, but it will not identify important features, such as for example non-isotropic autocorrelation, which led to that approximation.
%The utility of a non-\gls{ML} mathematical procedure over \gls{ML} is that the underlying theory, relations and mathematics are accessible to the user, which can be used to interpret the results.

%--------------------Paper contribution--------------------------
In this paper, we develop a data-driven model that generates approximate synthetic wall roughness for the purpose of grid-generation in numerical simulations. We use data-extraction tools to generate a two-dimensional array of the wall roughness height, starting from a single electron microscope image. This makes the present model more accessible than \gls{ML} models for CFD researchers, as a single roughness image can easily be extracted from literature, whereas a dataset of over twenty such images is unlikely to exist for the specific \gls{AM} roughness the CFD researcher may be interested in. 

The model combines the theory of Fourier analysis and the generation of synthetic fields inspired from an approach that Rogallo \cite{Rogallo1981} proposed to initialize homogeneous turbulence simulations with a random, correlated velocity field according to a prescribed energy spectrum. The surface height, as a function of surface coordinates, is approximated by a truncated \gls{FS}. The energy spectrum of the \gls{FS} is used to generate a synthetic surface roughness representation for a large surface that can be used in grid generators. 

The resulting synthetically generated roughness fields closely approximate the roughness of the input electron microscope image. In addition to good qualitative resemblance, the models reproduce quantitative features such the non-isotropic nature of the roughness autocorrelation, with negative short-distance autocorrelation in the $x-$direction and positive short- and long-distance autocorrelation in the $y-$direction. While the method is motivated and tested on \gls{AM} surface roughness, it can be applied to other roughness types as well.

%\tbk{Add motivation arguments for the presented method presented by Guus during meeting 8/23/2023? In my own words:}
%\textcolor{red}{Creating arbitrary size synthetic roughness surfaces from very little sample data of actual \gls{AM} roughness scans is the presented method's greatest strength, showing the importance of Fourier Analysis. The synthetic roughness field are generated using theory from the field of Turbulence, namely the Rogallo method, which generates homogeneous turbulence fields from an input Energy Spectrum and a random number generator. Connecting the principle of fluid flow to \gls{AM} roughness. \gls{AM} roughness in itself is a direct result of fluid flow, in the form of molten metal particles flowing to form a geometry during the \gls{SLM} process. There must thus be a direct relation between the \gls{AM} roughness and the air flow directly above the roughness. Making Turbulence theory a very important aspect of the presented method.}

%------------------Paper outline---------------------------------
A short review of Rogallo's method in 2D is given in \autoref{sec:RogalloReview}. In \autoref{sec:Methodology}, the methodology of the generation of the synthetic fields using Fourier analysis and Rogallo's method is described. Comparisons between the synthetic fields and the original \gls{AM} roughness are presented in \autoref{sec:Results}. Conclusions are made in \autoref{sec:Conclusions}. 
\section{Rogallo's Method Review}
\label{sec:RogalloReview}
Rogallo's method generates a model 3D homogeneous turbulence velocity field $\gls{sym:u}(\gls{sym:x}, \gls{sym:y}, \gls{sym:z})$ for the purpose of turbulence simulations, using a prescribed energy spectrum $\gls{sym:E}(\gls{sym:kvectormagnitude})$, where \gls{sym:kvectormagnitude} is the magnitude of the wave vector \gls{sym:kvector}. The turbulence field is generated in spectral space (indicated with the hat notation) using random number generators \cite{Rogallo1981}. 
%The presented method in this paper only requires Rogallo's method in 2D. Equations for $\gls{sym:alpha}$ and $\gls{sym:uhat}$ in 2D spectral space are derived in \autoref{sec:RandomNumberGenerator} and \ref{sec:SyntheticTurbulenceField} respectively, using the derivation presented by Rogallo.

The velocity covariance tensor in spectral space is defined as 

\begin{equation}
\label{eq:RelationRij}
    \gls{sym:Rtensorhat}[_{\gls{sub:ij}}] = \overline{\gls{sym:uhat}^*_{\gls{sub:j}}(\mathbf{k})\gls{sym:uhat}[_{\gls{sub:i}}](\mathbf{k})},
\end{equation}

where $\mathbf{k}$ is the wave number, $\gls{sym:uhat}[_{\gls{sub:i}}](\mathbf{k})$ is the velocity in Fourier space, * is used for complex conjugation and the overbar has the standard meaning of a Reynolds average.

Rogallo considers the relation between the energy spectrum and the trace of $\gls{sym:Rtensorhat}[_{\gls{sub:ij}}]$ \cite{NieuwstadtEtAl2016}:
\begin{equation}
\label{eq:RelationRijEk}
    E(|\mathbf{k}|) = \frac{1}{2} \varoiint_{|\mathbf{k}|} \hat{\mathbf{R}}_{ii}(\mathbf{k}) d\sigma,
\end{equation}
where $d\gls{sym:sigma}$ is an area differential on a 3D sphere of radius $|\mathbf{k}|$. The integral simplifies to the surface area of a sphere with radius \gls{sym:kvectormagnitude}. In 2D, integration takes place over circular rings, so the integral simplifies to the circumference of a ring with radius \gls{sym:kvectormagnitude}. The following condition is then obtained for the random number generator $\gls{sym:alpha}(\gls{sym:kvectormagnitude})$ to satisfy:
\begin{equation}
\label{eq:WeakRNGcondition}
    \hat{\mathbf{R}}_{ii}(k) = \frac{E(|\mathbf{k}|)}{\pi |\mathbf{k}|} = \overline{\alpha \alpha^*}.
\end{equation}

A form of \gls{sym:alpha} that satisfies the derived condition is: 
\begin{equation}
\label{eq:RandomNumberGenerator}
    \alpha \left(|\mathbf{k}|\right) = \sqrt{\frac{E\left(|\mathbf{k}|\right)}{\pi |\mathbf{k}|}} e^{\underline{i}\theta} \text{cos} \Phi,\\
\end{equation}
where \gls{sym:theta} and \gls{sym:Phi} are uniformly distributed random numbers in $[-\pi, \pi)$ and $[0, 2\pi)$ respectively and \gls{sym:imag} is the imaginary number.

An expression for the synthetic turbulence velocity field follows from continuity in spectral space: $\gls{sym:kvector}\cdot\hat{\mathbf{u}} = 0$, meaning that \gls{sym:kvector} and $\hat{\mathbf{u}}$ are orthogonal. We choose an orthonormal vector basis $\gls{sym:evector}[_{\gls{sub:n}\gls{sub:m}}] = \begin{bmatrix} \gls{sym:evector}[_{\gls{sub:n}}] & \gls{sym:evector}[_{\gls{sub:m}}] \end{bmatrix}^{\gls{sup:transpose}}$, with $\gls{sym:evector}[_{\gls{sub:m}}] = \begin{bmatrix}
    \frac{k_n}{\gls{sym:kvectormagnitude}} & \frac{k_m}{\gls{sym:kvectormagnitude}}
\end{bmatrix}^{\gls{sup:transpose}}$ and $\gls{sym:evector}[_{\gls{sub:n}}] = \begin{bmatrix}
    \frac{k_m}{\gls{sym:kvectormagnitude}} & -\frac{k_n}{\gls{sym:kvectormagnitude}}
\end{bmatrix}^{\gls{sup:transpose}}$, such that $\gls{sym:evector}[_{\gls{sub:m}}]$ is parallel to \gls{sym:kvector}. Then $\hat{\mathbf{u}}$ will only have a non-zero component along $\gls{sym:evector}[_{\gls{sub:n}}]$ to satisfy continuity: $\hat{\mathbf{u}}\left(k_n,k_m\right) = \gls{sym:alpha}(\gls{sym:kvectormagnitude})\gls{sym:evector}[_{\gls{sub:n}}] + 0\gls{sym:evector}[_{\gls{sub:m}}] $. Computations using $\hat{\mathbf{u}}$ are performed in spectral space, spanned by $k_n$ and $k_m$, resulting in the following expression:

\begin{equation}
\label{eq:SyntheticField2D}
    \hat{\mathbf{u}} \left(k_n, k_m\right) =  
    \begin{bmatrix}
        \alpha \left(|\mathbf{k}|\right) \frac{k_m}{|\mathbf{k}|} \\ - \alpha \left(|\mathbf{k}|\right) \frac{k_n}{|\mathbf{k}|}
    \end{bmatrix}.
\end{equation}

The same expression is obtained by simplifying Rogallo's 3D equations to 2D by removing the \gls{sym:z}-direction spectral wave number component terms. In the remainder of the paper, the notation $\gls{sym:u}$ is changed to $\gls{sym:f}$ since synthetic roughness fields are generated, not turbulence velocity fields. Also, the index notation is no longer used, $\gls{sub:i}$ and \gls{sub:j} are redefined later.
\section{Methodology}
\label{sec:Methodology}
The procedure for the generation of the synthetic \gls{AM} roughness is split up into three blocks: 1. Data Extraction, 2. Fourier Analysis and 3. Synthetic Field Generation. 

\begin{figure}[h!]
    \centering
    \includegraphics[width=0.8\textwidth]{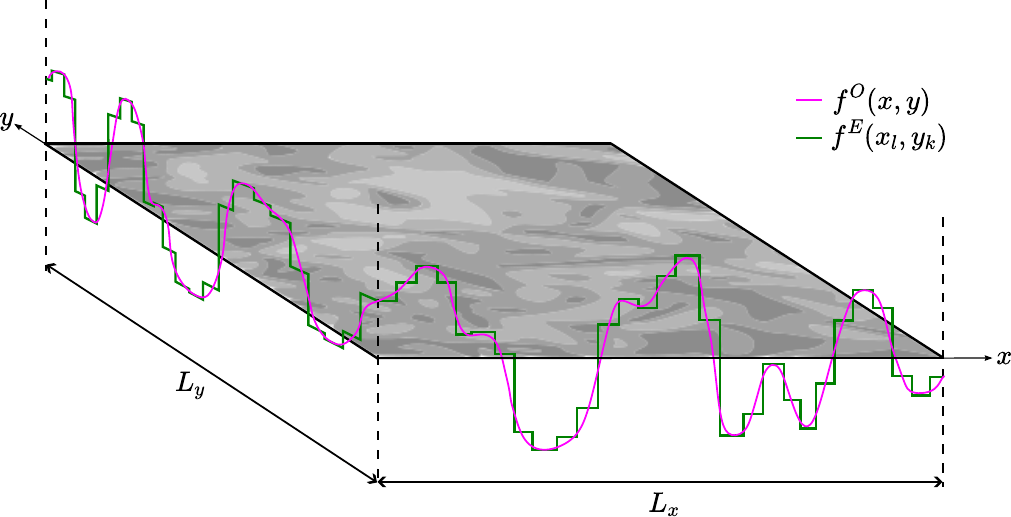}
    \caption{A schematic of the original and extracted image functions $\gls{sym:f}[^{\gls{sup:O}}]$ and $\gls{sym:f}[^{\gls{sup:E}}_{\gls{sub:l}\gls{sub:k}}]$.}
    \label{fig:ExtractedFunction}
\end{figure}

%____________________________________________________________________
\underline{\smash{\textbf{1. Data Extraction}}}\newline
The original roughness patch (superscript \gls{sup:O}) is represented by the function $\gls{sym:f}[^{\gls{sup:O}}] (\gls{sym:x}, \gls{sym:y})$ with roughness amplitudes that range from $\gls{sym:f}[^{\gls{sup:O}}_{\gls{sub:min}}]$ to $\gls{sym:f}[^{\gls{sup:O}}_{\gls{sub:max}}]$. 
For each pair of indices $l,k$, $f^{RGB}_{lk} = f^{RGB}(x_l, y_k)$ is an integer triple in $[0,255]\times[0,255]\times[0,255]$ which encodes the red, green and blue (RGB) color channels of the $lk-$th pixel of the electron microscope image. The RGB triples are extracted from the input image using
a three-step image data extraction procedure:
\begin{enumerate}
    \item \textbf{Importing the image}\newline
    The input image file is imported into the $\gls{sym:f}[^{\gls{sup:RGB}}_{\gls{sub:l}\gls{sub:k}}]$ array using the Pillow package in Python\footnote{Pillow, F. Lundh and A. Clark, 2011, \url{https://pillow.readthedocs.io/en/stable/index.html} [Accessed 11/14/2022]}.
    %, which converts each pixel of an image file to a list with three numbers and stores all pixels in an array. \newline
    
    \item \textbf{Converting the \gls{RGB} pixels to \gls{HSV}}:\newline
    The three \gls{RGB} values for each pixel in $\gls{sym:f}[^{\gls{sup:RGB}}_{lk}]$, $\gls{sym:R}[_{\gls{sub:l}\gls{sub:k}}]$, $\gls{sym:G}[_{\gls{sub:l}\gls{sub:k}}]$ and $\gls{sym:B}[_{\gls{sub:l}\gls{sub:k}}]$, are combined in a single hue value $\gls{sym:H}[_{\gls{sub:l}\gls{sub:k}}]$ in the \gls{HSV} color representation, via the following steps. 
    %Note that the substeps (ii) and (iii) are summarized in a function \texttt{rgb\_to\_hsv} from the \texttt{Colorsys} package in Python\footnote{Conversions between color systems, Python.org, \url{https://docs.python.org/3/library/colorsys.html} [Accessed 11/14/2022]}.
    \begin{enumerate}[label=(\roman*)]
        \item \textbf{Normalizing the \gls{RGB} representation}:\newline
        The \gls{RGB} values are divided by 255, yielding normalized $\gls{sym:Rprime}[_{\gls{sub:l}\gls{sub:k}}]$, $\gls{sym:Gprime}[_{\gls{sub:l}\gls{sub:k}}]$ and $\gls{sym:Bprime}[_{\gls{sub:l}\gls{sub:k}}] \in [0,1]$.
         
        \item \textbf{The conversion from \gls{RGB} to hue}:\newline 
        The conversion from \gls{RGB} to the hue \gls{sym:H} is defined by \autoref{eq:Hue} \cite{HanburyEtAl2002}. The three $\gls{sym:Rprime}[_{\gls{sub:l}\gls{sub:k}}]$, $\gls{sym:Gprime}[_{\gls{sub:l}\gls{sub:k}}]$ and $\gls{sym:Bprime}[_{\gls{sub:l}\gls{sub:k}}]$ are combined in a single value $\gls{sym:H}[_{\gls{sub:l}\gls{sub:k}}]$ in $[-1, 5]$.
        \begin{equation}
        \label{eq:Hue}
            H_{lk} = \begin{cases}
            \frac{G'_{lk} - B'_{lk}}{\text{max}(R'_{lk}, G'_{lk}, B'_{lk}) - \text{min}(R'_{lk}, G'_{lk}, B'_{lk})}, \qquad\qquad \!\! \text{if} \;\; R'_{lk} = \text{max}(R'_{lk}, G'_{lk}, B'_{lk})\\
            2 + \frac{B'_{lk} - R'_{lk}}{\text{max}(R'_{lk}, G'_{lk}, B'_{lk}) - \text{min}(R'_{lk}, G'_{lk}, B'_{lk})}, \qquad \text{if} \;\; G'_{lk} = \text{max}(R'_{lk}, G'_{lk}, B'_{lk})\\
            4 + \frac{R'_{lk} - G'_{lk}}{\text{max}(R'_{lk}, G'_{lk}, B'_{lk}) - \text{min}(R'_{lk}, G'_{lk}, B'_{lk})}, \qquad \text{if} \;\; B'_{lk} = \text{max}(R'_{lk}, G'_{lk}, B'_{lk})\\
            \end{cases}
        \end{equation}
        
        \item \textbf{Normalizing the hue}:\newline
        $\gls{sym:H}[_{\gls{sub:l} \gls{sub:k}}]$ is normalized to obtain $\gls{sym:Hprime}_{\gls{sub:l} \gls{sub:k}}\in[0,1]$
        \begin{equation}
        \label{eq:Hnormalization}
            H'_{lk} = \text{mod}_1\left(\frac{H_{lk}}{6}\right),
        \end{equation}
        where $\text{mod}_1$ stands for a modulo 1 operation
    \end{enumerate} 
    %The difficulty in terms of data extraction for images that have are displayed in terms of the \gls{HSV} colormap is the distinction between valleys and peaks, since both valleys and peaks are represented by the color red in the \gls{HSV} color scale as can be seen in \autoref{fig:fexact}. The key to distinguishing between valleys and peaks in the \gls{HSV} color map scale is the fact that the color red that represents peaks contains a small amount of blue: $R = 255$, $G = 0$, $B \neq 0$, while the color read that represents valleys is purely red: $R = 255$, $G = 0$, $B = 0$. \autoref{eq:Hue} converts the valley color red into zero, while it converts the peak color red into a number between 0 and -1. \autoref{eq:Hnormalization} the converts the peak value between 0 and -1 to a number that is closer to 1, while the valley 0 stays a zero. \newline
    
        \item \textbf{Scaling the \gls{sym:Hprime} to the original image scale}: \newline
    The last step of the image data extraction process involves the scaling of $\gls{sym:Hprime}[_{\gls{sub:l}\gls{sub:k}}]$ back to the range $[\gls{sym:f}[^{\gls{sup:O}}_{\gls{sub:min}}]$, $\gls{sym:f}[^{\gls{sup:O}}_{\gls{sub:max}}]]$ to yield the extracted (superscript \gls{sup:E}) function $\gls{sym:f}[^{\gls{sup:E}}_{\gls{sub:l}\gls{sub:k}}] = \gls{sym:f}[^{\gls{sup:E}}] (\gls{sym:x}[_{\gls{sub:l}}], \gls{sym:y}[_{\gls{sub:k}}])$: 
    \begin{equation}
    \label{eq:Hrescaling}
        f^{E}_{lk} = \left(f^{O}_{max} - f^{O}_{min}\right) H'_{lk} + f^{O}_{min}
    \end{equation}
\end{enumerate}
This image data extraction procedure is specific to input images with a \gls{HSV} colormap only, but is not difficult to adapt to other colormaps, provided the colormap definition is known. A schematic of the original and extracted functions $\gls{sym:f}[^{\gls{sup:O}}]$ and $\gls{sym:f}[^{\gls{sup:E}}_{\gls{sub:l}\gls{sub:k}}]$ is shown in \autoref{fig:ExtractedFunction}. For figure clarity, both functions are only shown on the \gls{sym:x} and \gls{sym:y}-axis, but span the entirety of the grey region. 

%____________________________________________________________________
\underline{\smash{\textbf{2. Fourier Analysis}}}\newline 
The Fourier modes are extracted from samples of $f^E$ at $x_i = i \Delta x$ and $y_j = j \Delta y$, where $i = 0, 1, 2, ..., N-1$ and $j = 0, 1, 2, ..., M-1$. $\Delta \gls{sym:x}$ and $\Delta \gls{sym:y}$ are different depending on if periodic and non-periodic samples are taken.

The function $\gls{sym:f}[^{\gls{sup:E}}_{\gls{sub:l}\gls{sub:k}}]$ is known at the discrete locations $\gls{sym:x}[_{\gls{sub:l}}]$ and $\gls{sym:y}[_{\gls{sub:k}}]$. Cubic interpolation (superscript \gls{sup:I}) is used to allow for sampling of $\gls{sym:f}[^{\gls{sup:E}}_{\gls{sub:l}\gls{sub:k}}]$ at any location other then $\gls{sym:x}[_{\gls{sub:l}}]$ and $\gls{sym:y}[_{\gls{sub:k}}]$. The result is the continuous function $\gls{sym:f}[^{\gls{sup:I}}]\left(\gls{sym:x},\gls{sym:y}\right)$. Taking samples at $\gls{sym:x}[_{\gls{sub:i}}]$ and $\gls{sym:y}[_{\gls{sub:j}}]$ results in the discrete sample function $\gls{sym:f}[^{\gls{sup:I}}_{ij}] = \gls{sym:f}[^{\gls{sup:I}}]\left(\gls{sym:x}[_{\gls{sub:i}}],\gls{sym:y}[_{\gls{sub:j}}]\right)$.

%If $\gls{sym:f}[^{\gls{sup:I}}]$ is periodic, the samples on the $\gls{sym:x}=0$ axis are the same as the samples located at $\gls{sym:x} = \gls{sym:L}[_{\gls{sub:x}}]$. The same is true for the samples in the $\gls{sym:y}$-direction at $\gls{sym:y}=0$ and $\gls{sym:y} = \gls{sym:L}[_{\gls{sub:y}}]$. The \gls{sym:N} and \gls{sym:M} samples in \gls{sym:x} and \gls{sym:y} respectively are therefore taken such that the locations $\gls{sym:x} = \gls{sym:L}[_{\gls{sub:x}}]$ and $\gls{sym:y} = \gls{sym:L}[_{\gls{sub:y}}]$ are omitted. The distance between periodic samples therefore equals: $\Delta x = L_x/N$ and $\Delta y = L_y/M$.

%Because it can be expected that values obtained from a scanned image are not exactly periodic, with the samples at $\gls{sym:x}=0$ and $\gls{sym:y}=0$ are not being the same at $\gls{sym:x} = \gls{sym:L}[_{\gls{sub:x}}]$ and $\gls{sym:y} = \gls{sym:L}[_{\gls{sub:y}}]$, the function $\gls{sym:f}[^{\gls{sup:I}}]$ can also be sampled \gls{sym:N} and \gls{sym:M} times in the \gls{sym:x} and \gls{sym:y}-direction respectively, including the domain boundaries at $\gls{sym:x} = \gls{sym:L}[_{\gls{sub:x}}]$ and $\gls{sym:y} = \gls{sym:L}[_{\gls{sub:y}}]$. The distance between the samples is given by: $\Delta x = L_x/(N-1)$ and $\Delta y = L_y/(M-1)$.

%\subsection{Fourier Analysis}
%\label{sec:FourierAnalysis}
The modes inside $\gls{sym:f}[^{\gls{sup:I}}_{\gls{sub:i}\gls{sub:j}}]$ are extracted by transforming the samples to spectral space using the \gls{DFT} in 2D as given by \autoref{eq:DFTfIij}, yielding the Fourier (superscript \gls{sup:F}) coefficients $\gls{sym:fhat}[^{\gls{sup:F}}_{nm}] = \gls{sym:fhat}[^{\gls{sup:F}}]\left(k_n, k_m\right)$ of $\gls{sym:f}[^{\gls{sup:O}}]$:  
\begin{equation}
\label{eq:DFTfIij}
    \hat{f}^{F}_{nm} = \frac{1}{N M} \sum^{N-1}_{i = 0} \sum^{M-1}_{j = 0} f^{I}_{ij} e^{-k_n x_i \underline{i}} e^{-k_m y_j \underline{i}},
\end{equation}

where $\gls{sym:k}[_{\gls{sub:n}}] = 2\pi\gls{sym:n}/\gls{sym:L}[_{\gls{sub:x}}]$ and $
\gls{sym:k}[_{\gls{sub:m}}] = 2\pi\gls{sym:m}/\gls{sym:L}[_{\gls{sub:y}}]$, with $\gls{sym:n} = -\gls{sym:N}/2+1, ..., \gls{sym:N}/2$ and $\gls{sym:m} = -\gls{sym:M}/2+1, ..., \gls{sym:M}/2$.

Using the modes $\gls{sym:fhat}[^{\gls{sup:F}}_{\gls{sub:n}\gls{sub:m}}]$ of the original function $\gls{sym:f}[^{\gls{sup:O}}]$, a \gls{FS} representation $\gls{sym:f}[^{\gls{sup:F}}]\left(\gls{sym:x}, \gls{sym:y}\right)$ in physical space can be obtained using the \gls{IDFT} in 2D:
\begin{equation}
\label{eq:IDFT}
    f^F\left(x, y\right) = \sum^{N/2}_{n = -N/2 + 1} \sum^{M/2}_{m = -M/2 + 1} \hat{f}^F_{nm} e^{k_n x \underline{i}} e^{k_m y \underline{i}}
\end{equation}

%\subsection{Roughness Field Energy Spectrum}
%\label{sec:EkPreRogallo}
To apply Rogallo's method as described above, the energy spectrum of the Fourier coefficients, $\gls{sym:E}[^{\gls{sup:F}}]\left(\gls{sym:kvectormagnitude}\right)$, is obtained by integrating over rings with radius \gls{sym:kvectormagnitude} in spectral space and normalizing using the ring circumference, as given by \autoref{eq:FourierEnergy}, with $\gls{sym:kvectormagnitude} \in \left[0,\gls{sym:kvectormagnitude}[_{\gls{sub:max}}]\right]$, where $\gls{sym:kvectormagnitude}[_{\gls{sub:max}}] = \sqrt{\left(\frac{N}{2}\right)^2 + \left(\frac{M}{2}\right)^2}$. For every ring of radius \gls{sym:kvectormagnitude}, the energy is binned to the nearest integer value of \gls{sym:kvectormagnitude} to obtain the complete energy spectrum:

\begin{equation}
\label{eq:FourierEnergy}
        E\left(|\mathbf{k}|\right)  = \oint \left(\hat{f}_{nm}\right)^2 d \sigma \approx \sum\limits_{|\mathbf{k}|-0.5 \leq \sqrt{k_n^2 + k_m^2}  < |\mathbf{k}|+ 0.5} \left(\hat{f}_{nm}\right)^2.
\end{equation}

%\begin{equation}
%\label{eq:FourierEnergy}
%       E\left(|\mathbf{k}|\right)  = \oint \frac{\left(\hat{f}_{nm}\right)^2}{2 \pi |\mathbf{k}|} d \sigma,
%\end{equation}

%with $\hat{f}_{nm}(k_1,k_2) = \hat{f}^F_{nm} (\lfloor k_1 \rceil, \lfloor k_2 \rceil)$, with $\lfloor \cdot \rceil$ denoting rounding to the nearest multiple of $2\pi/L_x, 2\pi /L_y$ respectively.
%\tbk{I think the previous sentence should be something in the lines of:}\newline
%with $\hat{f}_{nm} = \hat{f}^F_{nm}$ for Rogallo's method, but $\hat{f}_{nm}$ can be taken as any other spectral field in order to extract its energy spectrum. Integrating along 2D spectral rings, the energy integral definition simplifies to $\gls{sym:E}(\gls{sym:kvectormagnitude}) = \gls{sym:fhat}[^{2}_{\gls{sub:n}\gls{sub:m}}]$, where $\gls{sym:E}(\gls{sym:kvectormagnitude})$ at every discrete location $\gls{sym:kvector}[_{\gls{sub:n}}]$, $\gls{sym:kvector}[_{\gls{sub:m}}]$ is binned to $\lfloor \gls{sym:kvectormagnitude} \rceil$, with $\lfloor \cdot \rceil$ denoting the nearest rounded integer value of \gls{sym:kvectormagnitude}. 

\underline{\smash{\textbf{3. Syntetic Field Generation}}}\newline
The synthetic roughness fields are generated using the method by Rogallo \cite{Rogallo1981} in 2D, as summarized in \autoref{sec:RogalloReview}. The Rogallo (superscript \gls{sup:R}) spectral synthetic vector roughness field $\gls{sym:fhat}[^{\gls{sup:R}}_{\gls{sub:n}\gls{sub:m}}]$ is given by \autoref{eq:SpectralSyntheticField}. The vector field components of $\gls{sym:fhat}[^{\gls{sup:R}}_{\gls{sub:n}\gls{sub:m}}]$, $\gls{sym:fhat}[^{\gls{sup:R}}_{\gls{sub:n}}]$ and $\gls{sym:fhat}[^{\gls{sup:R}}_{\gls{sub:m}}]$, are given by \autoref{eq:SpectralSyntheticRoughnesFieldn} and \ref{eq:SpectralSyntheticRoughnesFieldm} respectively. The random number generator \gls{sym:alpha} is described by \autoref{eq:RandomNumberGeneratorEF}.
\begin{equation}
\label{eq:SpectralSyntheticField}
    \hat{f}^{R}_{nm} = \begin{bmatrix}
        \hat{f}^{R}_{n} \\ \hat{f}^{R}_{m}
    \end{bmatrix}
\end{equation}
\begin{equation}
\label{eq:SpectralSyntheticRoughnesFieldn}
    \hat{f}^{R}_{n} = \alpha \left(|\mathbf{k}|\right) \frac{k_m}{|\mathbf{k}|}
\end{equation}
\begin{equation}
\label{eq:SpectralSyntheticRoughnesFieldm}
    \hat{f}^{R}_{m} = -\alpha \left(|\mathbf{k}|\right) \frac{k_n}{|\mathbf{k}|}
\end{equation}
\begin{equation}
\label{eq:RandomNumberGeneratorEF}
    \alpha (|\mathbf{k}|) = \sqrt{\frac{E^F (|\mathbf{k}|)}{\pi |\mathbf{k}|}} e^{\underline{i} \theta} \text{cos}(\Phi),
\end{equation}

with $\theta \sim U(-\pi,\pi), \Phi \sim U(0,2\pi)$.

The spectral synthetic roughness fields are transformed back to physical space using the \gls{IDFT}, \autoref{eq:IDFT}, to obtain the physical synthetic roughness fields. We note that in the above procedure $E^F(|\mathbf{k}|)$ is uniquely detemined by the input image, but $\alpha$ is a random variable. Thus, Rogallo's method enables the generation of an unlimited amount of synthetic roughness fields from a single input scan. 

The energy spectrum used to generate the synthetic roughness is not an error-less representation of the energy spectrum of the input image scan, since the image data extraction and the interpolation part of the procedure used to extract the energy spectrum introduce errors. Additionally, the energy spectrum of the synthetic field will contain sampling errors, since $| \alpha |^2$ is equal to $\frac{E^F(| \mathbf{k}|)}{2\pi | \mathbf{k}|}$ in expectation only. These sampling errors can be reduced by increasing $N$ and $M$.

%\subsection{Scaling}
%\label{sec:Scaling}
Without loss of generality, we set $\gls{sym:L}[_{\gls{sub:x}}] = \gls{sym:L}[_{\gls{sub:y}}] = 2\pi$, so that the wave numbers $k_n$ and $k_m$ have the same values as \gls{sym:n} and \gls{sym:m} respectively. The roughness fields can then be scaled to any desired size. The scaled domain lengths in \gls{sym:x} and \gls{sym:y} are referred to as $\gls{sym:Ltilde}[_{\gls{sub:x}}]$ and $\gls{sym:Ltilde}[_{\gls{sub:y}}]$ respectively, which are obtained by dividing $\gls{sym:L}[_{\gls{sub:x}}]$ and $\gls{sym:L}[_{\gls{sub:y}}]$ by the scaling factors $s = 2\pi / \tilde{L}_x$ and $r = 2\pi / \tilde{L}_y$ respectively. In spectral space, the scaled wave numbers are given by $
\tilde{k}_n = 2\pi\gls{sym:n}/\gls{sym:Ltilde}[_{\gls{sub:x}}]$ and $
\tilde{k}_m = 2\pi\gls{sym:m}/\gls{sym:Ltilde}[_{\gls{sub:y}}]$.

\section{Tests}
\label{sec:Results}
\begin{figure}[h!]
    \centering
    \includegraphics[width=0.9\textwidth]{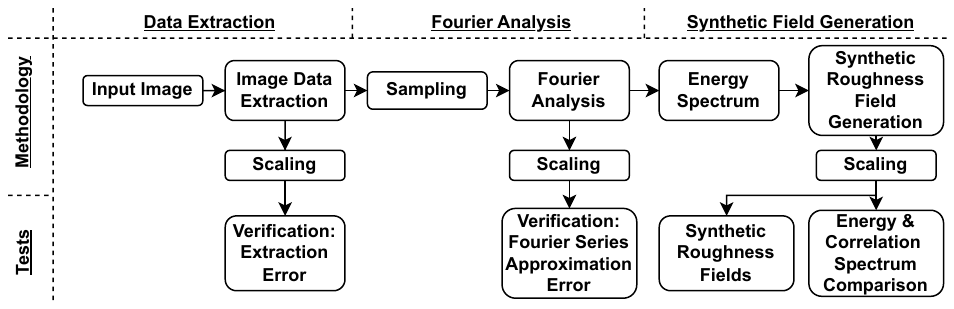}
    \caption{The structure of the synthetic field generation methodology and its results.}
    \label{fig:MethodStructure}
\end{figure}

The synthetic roughness generation procedure is summarized in \autoref{fig:MethodStructure}. Various tests are performed using the synthetic roughness generation procedure, which are discussed in their respective subsections.

\subsection{Verification of Image Extraction Accuracy}
\label{eq:ExtractionErrorMethod}
To test the accuracy of the image extraction procedure, we consider a test function $\gls{sym:f}^{\gls{sup:O}}(\gls{sym:x},\gls{sym:y}) = \text{sin}(x) + \text{cos}(2y)$ on the domain of $\gls{sym:L}[_{\gls{sub:x}}], \gls{sym:L}[_{\gls{sub:y}}] = 2 \pi$ and define the extraction error as $\varepsilon^E = \dfrac{\left\Vert f_{lk} - f^O  \right\Vert_\infty}{\left\Vert f^O  \right\Vert_\infty}$.
The extraction error is calculated as $6\%$ on a low resolution $368\times369$ \gls{HSV} colormap image; on a high-resolution $1486\times1486$ image, the error is $2\%$. The extraction error originates from the difference between the continuous and discrete function $\gls{sym:f}[^{\gls{sup:O}}]$ and $\gls{sym:f}[^{\gls{sup:E}}_{\gls{sub:l}\gls{sub:k}}]$, see the schematic in \autoref{fig:ExtractedFunction}.

\subsection{Verification of Fourier Series Approximation Error}
\label{sec:FSErrorMethod}

%The synthetic roughness fields are generated using the energy spectrum of Fourier coefficient extracted from an input scan. However, the Fourier coefficients are not extracted with zero error, meaning that there is an error between the extracted and actual Fourier coefficients of the input image. The total error in the extracted Fourier coefficients consists of the image extraction and interpolation error. 

The $L^2$ \gls{FS} approximation error is defined as $\varepsilon^F = \frac{\left\Vert f_{lk}^F - f^E_{lk}  \right\Vert_2}{\left\Vert f^O  \right\Vert_2}$. For $\gls{sym:N}, \gls{sym:M} = 64$ (low sample quantity) with periodic sampling, $\varepsilon^F=1.11\%$. For $\gls{sym:N}, \gls{sym:M} = 256$ (high sample quantity) with periodic sampling, $\varepsilon^F=0.25\%$ and for $\gls{sym:N}, \gls{sym:M} = 256$ with non-periodic sampling, $\varepsilon^F=0.19\%$. Based on this, we conclude that $\gls{sym:N}, \gls{sym:M} = 64$ is sufficient for the purpose of synthetic roughness generation.

\subsection{Synthetic Roughness Fields}
\label{sec:SyntheticRoughnessFields}

\begin{figure}[h!]
    \centering
    \includegraphics[width=0.8\textwidth]{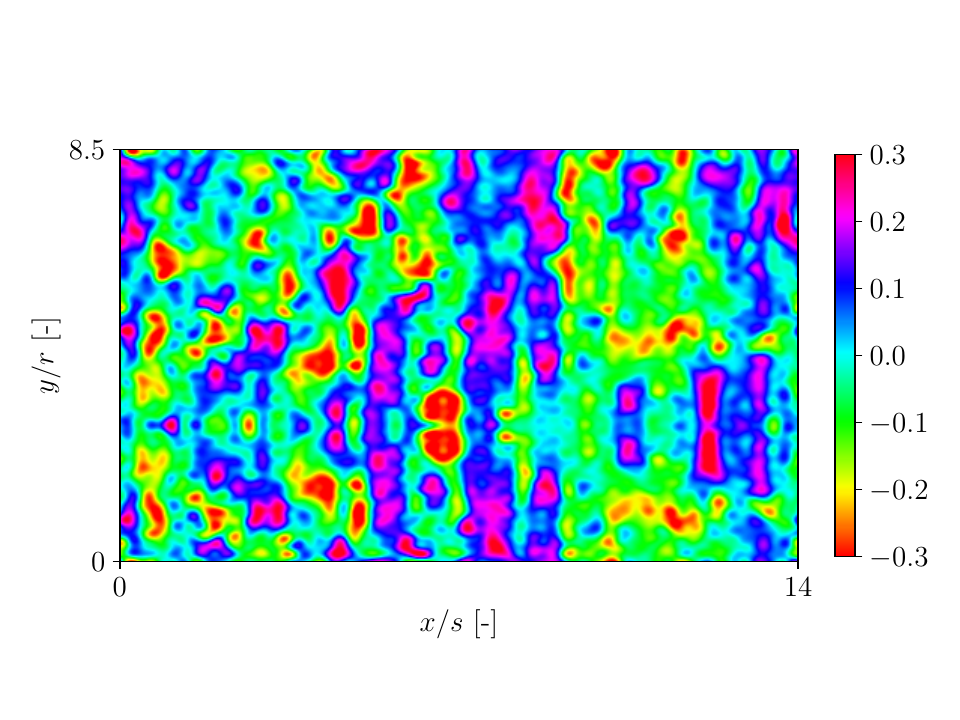}
    \caption{The Fourier-Series approximation $\gls{sym:f}[^{\gls{sup:F}}]$ of the input roughness surface by Altland \textit{et al.} \cite{AltlandEtAl2022} created with \gls{sym:N},\gls{sym:M} = 64 period samples, plotted with a resolution of $\gls{sym:N}[_{\gls{sub:plot}}] = 1950$, $\gls{sym:M}[_{\gls{sub:plot}}] = 1186$ and scaled with the scaling factors $\gls{sym:s} = 2\pi/14$ and $\gls{sym:r} = 2\pi/8.5$.}
    \label{fig:fF}
\end{figure}

An \gls{AM} roughness scan image by Altland \textit{et al.}\cite{AltlandEtAl2022} (identified as S3 in that work) is used as the input image. The \gls{AM} input image, shown on \autoref{fig:fF} has a resolution of $1950\times1186$ pixels, from which $\gls{sym:N}, \gls{sym:M} = 64$ periodic samples are taken.

The output of Rogallo's method is a two-dimensional vector field, based on which a scalar roughness field is extracted. We test several alternative definitions of this scalar field as a function of the Rogallo vector field. The $x-$ and $y-$components of the vector field, denoted as $f_x^R$ and $f_y^R$ are natural choices, as is its magnitude, $f^R$. Note that the magnitude is a strictly positive scalar, whereas roughness topography has both positive and negative peaks. On the other hand, vorticity is symmetric with respect to $0$, uses both components of the Rogallo vector field, and highlights the Rogallo field's structure, as it does in turbulence\cite{NieuwstadtEtAl2016}. Thus, two additional scalar fields which will be evaluated here are the vorticity of the Rogallo vector field, $f^V$, as well as its enstrophy, $f^{V^2}$, defined as the square of the vorticity.

The synthetic fields are scaled back to the domain aspect ratio and the roughness amplitude range of the original \gls{AM} input image.
Furthermore, all synthetic fields in this section are plotted using a discrete resolution according to the variables $\gls{sym:x}[_{\gls{sub:plot}}] \in \left[0, L_x\right]$ and $\gls{sym:y}[_{\gls{sub:plot}}] \in \left[0, L_y\right]$ with at a resolution of $\gls{sym:N}[_{\gls{sub:plot}}]$ and $\gls{sym:M}[_{\gls{sub:plot}}]$ in the \gls{sym:x} and \gls{sym:y} direction respectively. All fields resulting from the \gls{IDFT} are plotted in the form:
\begin{equation*}
\label{eq:FourierSeries}
    f\left(x_{plot}, y_{plot}\right) = \sum^{N/2}_{n = -N/2 + 1} \sum^{M/2}_{m = -M/2 + 1} \hat{f}_{nm} e^{k_n x_{plot} \underline{i}} e^{k_m y_{plot} \underline{i}}.
\end{equation*}

\begin{figure}[h!]
  \centering
    \includegraphics[width=0.8\textwidth]{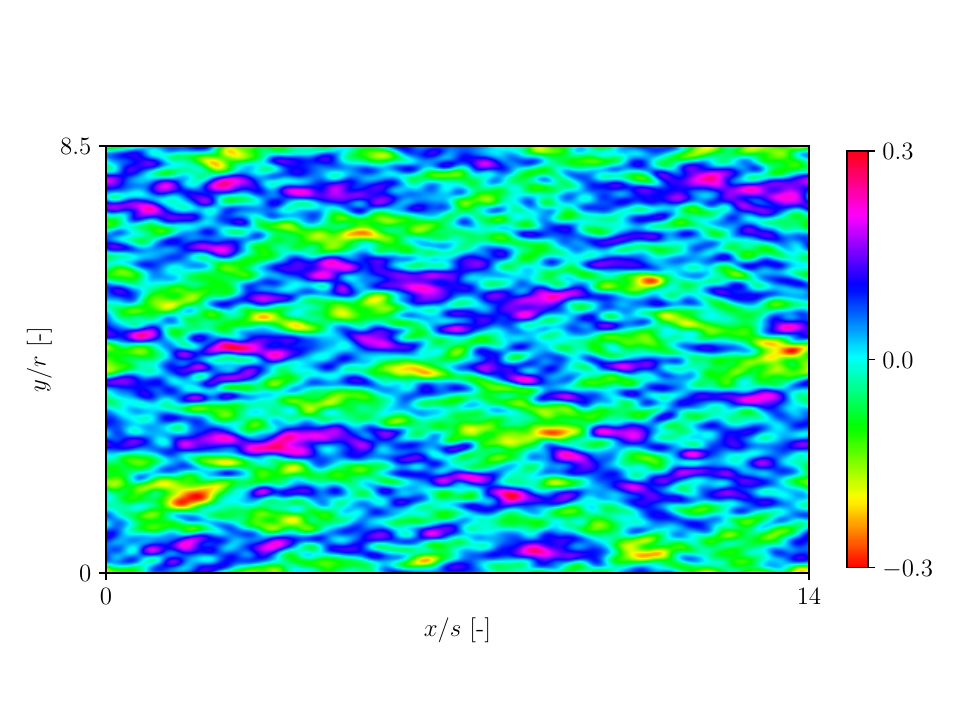}
    \caption{The \gls{sym:x}-vector component synthetic roughness field $\gls{sym:ftilde}[^{\gls{sup:R}}_{\gls{sub:x}}]$, plotted with a resolution of $\gls{sym:N}[_{\gls{sub:plot}}]$, $\gls{sym:M}[_{\gls{sub:plot}}]$ = 500 and the domain scaling factors $\gls{sym:s} = 2\pi/14$ and $\gls{sym:r} = 2\pi/8.5$.}
    \label{fig:fRx}
\end{figure}

\begin{figure}
    \centering
    \includegraphics[width=0.8\textwidth]{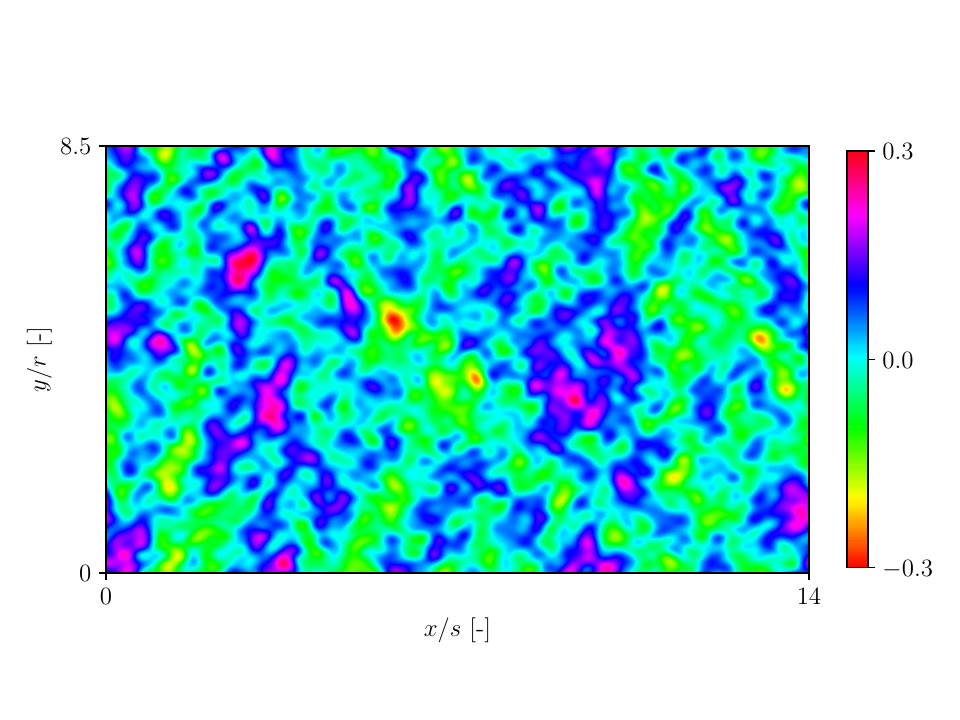}
    \caption{The \gls{sym:y}-vector component synthetic roughness field $\gls{sym:ftilde}[^{\gls{sup:R}}_{\gls{sub:y}}]$, plotted with a resolution of $\gls{sym:N}[_{\gls{sub:plot}}]$, $\gls{sym:M}[_{\gls{sub:plot}}]$ = 500 and the scaling factors $\gls{sym:s} = 2\pi/14$ and $\gls{sym:r} = 2\pi/8.5$.}
    \label{fig:fRy}
\end{figure}

\begin{figure}[h!]
    \centering
    \includegraphics[width=0.8\textwidth]{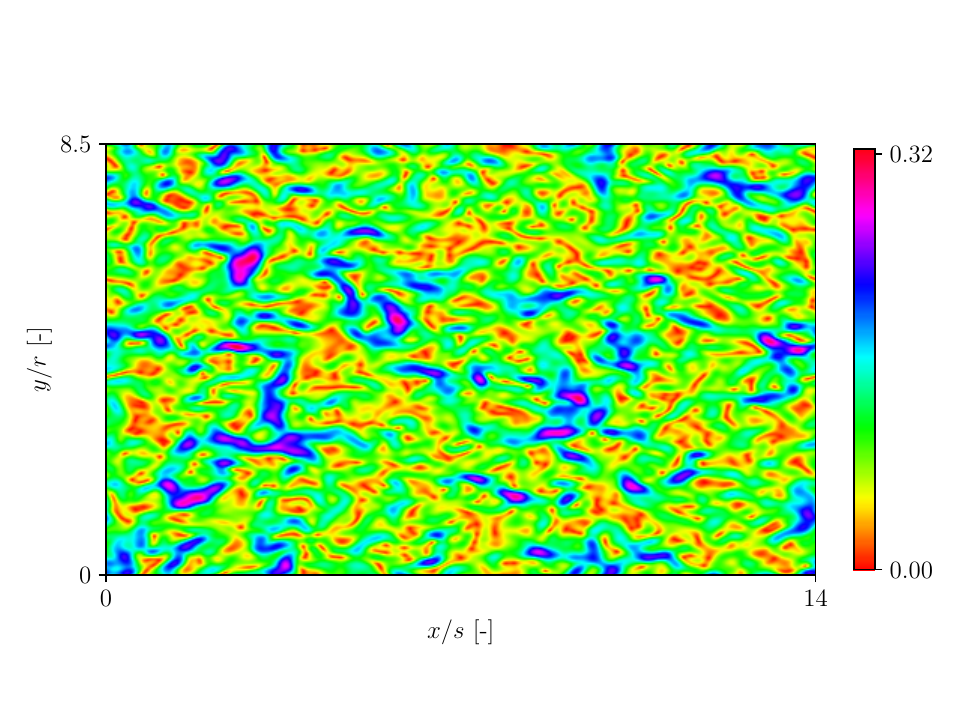}
    \caption{The magnitude synthetic roughness field $\gls{sym:ftilde}[^{\gls{sup:R}}]$, plotted with a resolution of $\gls{sym:N}[_{\gls{sub:plot}}]$, $\gls{sym:M}[_{\gls{sub:plot}}] = 500$ and the scaling factors $\gls{sym:s} = 2\pi/14$ and $\gls{sym:r} = 2\pi/8.5$.}
    \label{fig:fR}
\end{figure}

The \gls{FS} approximation of the \gls{AM} input is presented in \autoref{fig:fF}. The sythetic fields $f_x^R$, $f_y^R$ and $f^R$ are presented in \autoref{fig:fRx}, \ref{fig:fRy} and \ref{fig:fR} respectively.

\begin{figure}[h!]
  \centering
    \includegraphics[width=0.8\textwidth]{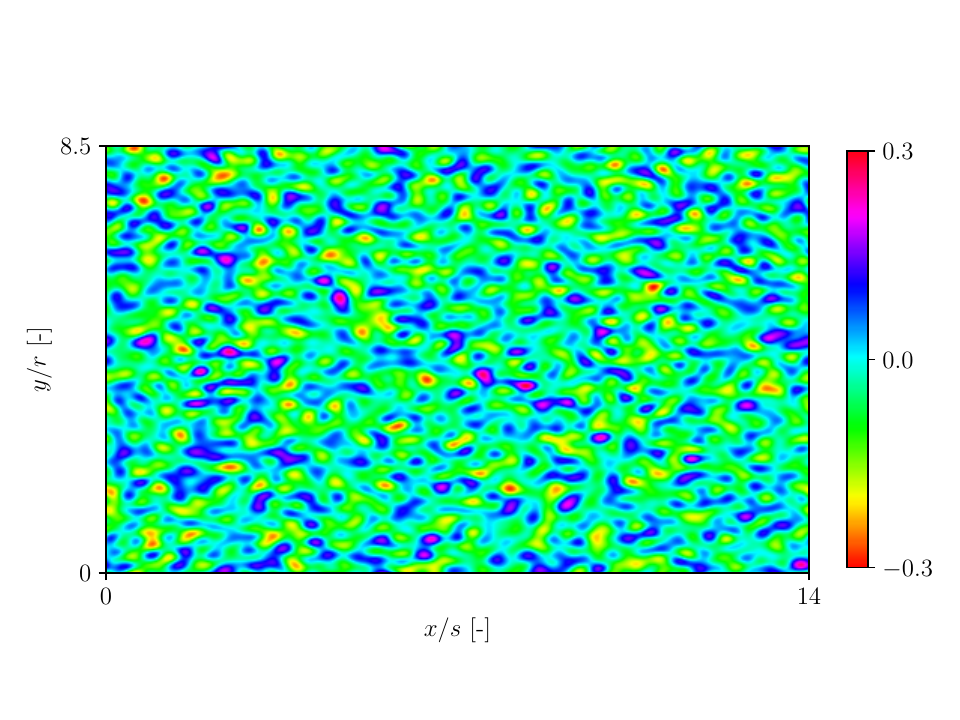}
    \caption{The filtered synthetic vorticity field $\gls{sym:ftilde}[^{\gls{sup:V}}]$, plotted with a resolution of $\gls{sym:N}[_{\gls{sub:plot}}]$, $\gls{sym:M}[_{\gls{sub:plot}}]$ = 500 and the scaling factors $\gls{sym:s} = 2\pi/14$ and $\gls{sym:r} = 2\pi/8.5$.}
    \label{fig:fV}
\end{figure}

\begin{figure}[h!]
    \centering
    \includegraphics[width=0.8\textwidth]{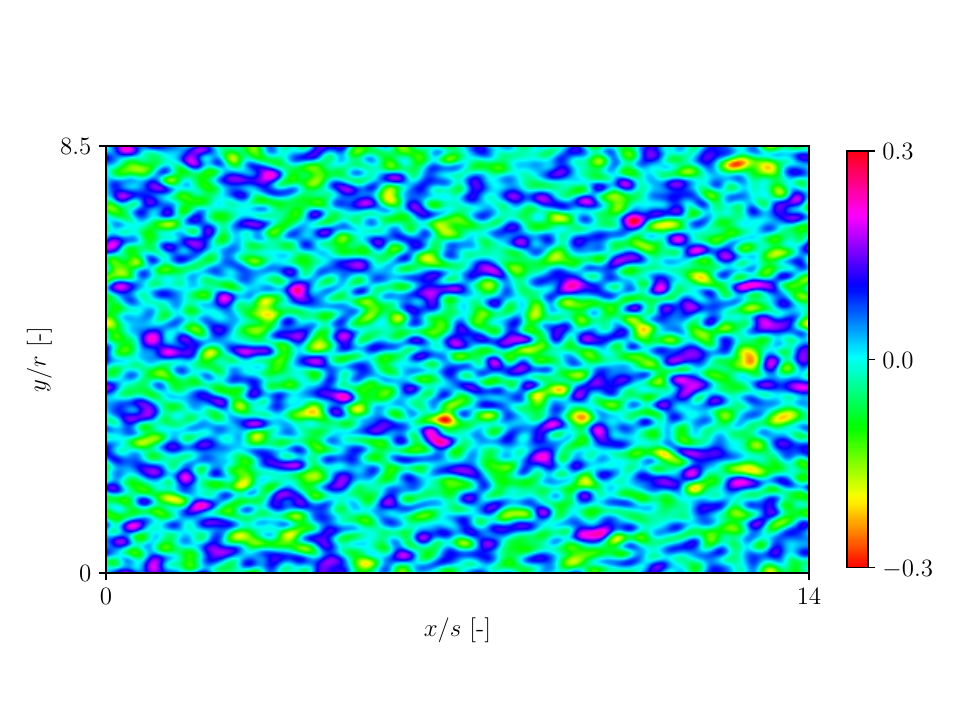}
    \caption{The synthetic enstrophy field $\gls{sym:ftilde}[^{\gls{sup:V2}}]$, plotted with a resolution of $\gls{sym:N}[_{\gls{sub:plot}}]$, $\gls{sym:M}[_{\gls{sub:plot}}]$ = 500 and the scaling factors $\gls{sym:s} = 2\pi/14$ and $\gls{sym:r} = 2\pi/8.5$.}
    \label{fig:fV2}
\end{figure}

We note that taking the vorticity of the Rogallo field skews its energy spectrum to the higher wavenumbers. To remove this undesired high-frequency content, a top hat filter is applied to the synthetically generated Fourier coefficients $\gls{sym:fhat}[^{\gls{sup:V}}_{\gls{sub:n}\gls{sub:m}}]$ that sets all $\gls{sym:fhat}[^{\gls{sup:V}}_{\gls{sub:n}\gls{sub:m}}]$ at $\gls{sym:kvectormagnitude} = \sqrt{k_n^2 + k_m^2} \geq 32$ equal to zero in order to remove the smallest scales. The enstrophy is, however, not filtered, as the squaring in it already serves to reduce the energy of the higher-frequency wavenumbers. The synthetic vorticity and enstrophy fields, $f^V$ and $f^{V^2}$, are presented in \autoref{fig:fV} and \ref{fig:fV2} respectively.

\subsection{Energy Spectrum Comparison}
\label{sec:EnergySpectrumComparison}
\begin{figure}[h!]
    \centering
    \includegraphics[width=0.8\textwidth]{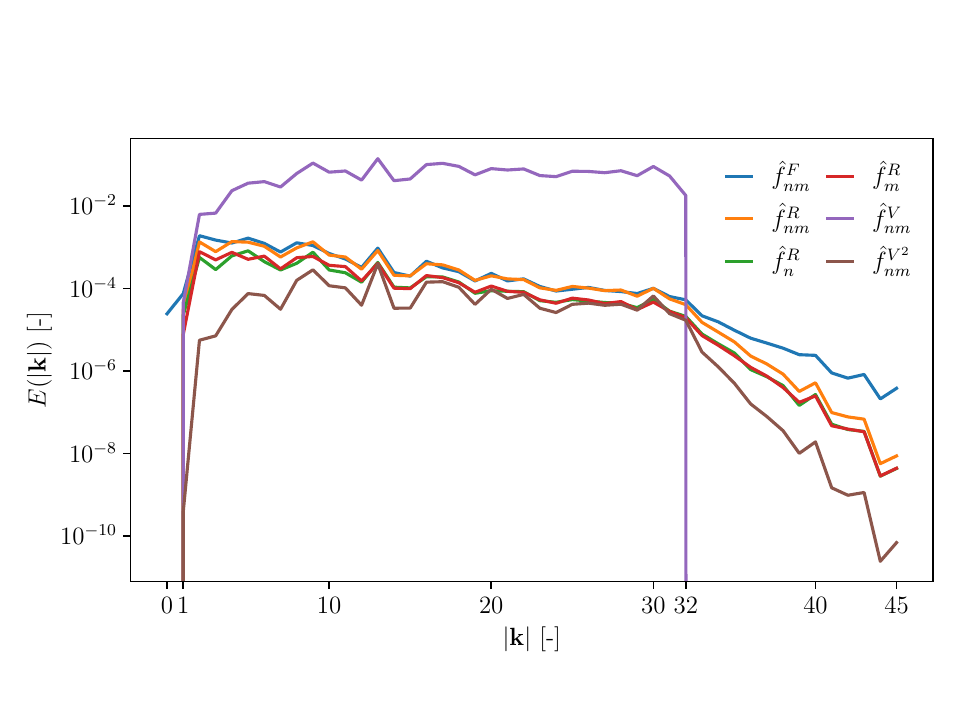}
    \caption{The energy spectrum $\gls{sym:E} (\gls{sym:kvectormagnitude})$ comparison of the AM surface \gls{FS} approximation and the five synthetic fields Fourier coefficients.}
    \label{fig:EnergySpectrumComparison}
\end{figure}

The energy spectrum is extracted from the Fourier coefficients of the \gls{AM} \gls{FS} approximation field $\gls{sym:fhat}[^{\gls{sup:F}}_{\gls{sub:n}\gls{sub:m}}]$ and the five synthetically generated roughness fields $\gls{sym:fhat}[^{\gls{sup:R}}_{\gls{sub:n}}]$, $\gls{sym:fhat}[^{\gls{sup:R}}_{\gls{sub:m}}]$, $\gls{sym:fhat}[^{\gls{sup:R}}_{\gls{sub:n}\gls{sub:m}}]$, $\gls{sym:fhat}[^{\gls{sup:V}}_{\gls{sub:n}\gls{sub:m}}]$, $\gls{sym:fhat}[^{\gls{sup:V2}}_{\gls{sub:n}\gls{sub:m}}]$, using the methods described in \autoref{sec:Methodology} with \gls{sym:N}, \gls{sym:M} = 64 periodic samples. The energy spectra are presented in \autoref{fig:EnergySpectrumComparison}.
The energy spectrum of the Fourier coefficients $\gls{sym:fhat}[^{\gls{sup:F}}_{\gls{sub:n}\gls{sub:m}}]$ represents the original \gls{AM} surface input image. The energy spectrum of $\gls{sym:fhat}[^{\gls{sup:R}}_{\gls{sub:n}\gls{sub:m}}]$, closely matches that of $\gls{sym:fhat}[^{\gls{sup:F}}_{\gls{sub:n}\gls{sub:m}}]$, which verifies the implementation of the method by Rogallo. The $\gls{sym:fhat}[^{\gls{sup:R}}_{\gls{sub:n}}]$ and $\gls{sym:fhat}[^{\gls{sup:R}}_{\gls{sub:m}}]$ synthetic fields contain less energy than $\gls{sym:fhat}[^{\gls{sup:R}}_{\gls{sub:n}\gls{sub:m}}]$, since they are the components of $\gls{sym:fhat}[^{\gls{sup:R}}_{\gls{sub:n}\gls{sub:m}}]$.  
Due to the multiplication by $\mathbf{k}$ in Fourier space when vorticity is evaluated, the synthetic vorticity field's spectral energy is one to two orders of magnitude higher than $\gls{sym:fhat}[^{\gls{sup:F}}_{\gls{sub:n}\gls{sub:m}}]$ and the other synthetic fields. Additionally, the vorticity field's energy spectrum is uniform across the range of $|\mathbf{k}|$, whereas the spectral energy of the Rogallo synthetic fields decreases by an order of magnitude from $|\mathbf{k}|=1$ to $|\mathbf{k}|=32$. The effect of the top hat filter removing all vorticity modes at $\gls{sym:kvectormagnitude} \geq 32$ is shown in $\gls{sym:E}[^{\gls{sup:V}}]$ by means of a sharp drop off in the energy at $\gls{sym:kvectormagnitude} = 32$. 
The synthetic enstrophy field Fourier coefficients $\gls{sym:fhat}[^{\gls{sup:V2}}_{\gls{sub:n}\gls{sub:m}}]$ contains a similar amount of energy as the synthetic components fields $\gls{sym:fhat}[^{\gls{sup:R}}_{\gls{sub:n}}]$ and $\gls{sym:fhat}[^{\gls{sup:R}}_{\gls{sub:m}}]$ for the middle range of wave numbers, but less energy in the lower wave numbers. Similarly to the synthetic vorticity field, the low and high wave number modes spectral energy is decreased and amplified respectively in the enstrophy field.
%In terms of the Gaussian roughness model, the energy present does not closely match the energy spectra of the \gls{FS} approximation or any of the synthetic fields.

At $\gls{sym:kvectormagnitude} \geq 32$, the energy spectra of $\gls{sym:fhat}[^{\gls{sup:F}}_{\gls{sub:n}\gls{sub:m}}]$ and $\gls{sym:fhat}[^{\gls{sup:R}}_{\gls{sub:n}\gls{sub:m}}]$ no longer match. At $|\mathbf{k}|=40$, the energy of $\gls{sym:fhat}[^{\gls{sup:R}}_{\gls{sub:n}\gls{sub:m}}]$ is lower by an order of magnitude compared to $\gls{sym:fhat}[^{\gls{sup:F}}_{\gls{sub:n}\gls{sub:m}}]$, as a result of the normalization in the energy spectrum definition of \autoref{eq:FourierEnergy}. At $\gls{sym:kvectormagnitude} \geq 32$ the rings used for energy extraction partially extend beyond the domain, only capturing the energy present in the corners of the domain. Even though the magnitude of $\gls{sym:E}$ is smaller, it is normalized with the circumference of the entire energy ring, resulting in an effectively lower normalized energy at $\gls{sym:kvectormagnitude} \geq 32$.

\subsection{Two-Point Correlation Spectra Comparison}
\label{sec:CorrelationSpectraComparison}
The two-point correlation spectra \cite{NieuwstadtEtAl2016} are extracted from the roughness surface \gls{FS} approximation $\gls{sym:f}[^{\gls{sup:F}}]$ and the five synthetic fields $\gls{sym:f}[^{\gls{sup:R}}_{\gls{sub:x}}]$, $\gls{sym:f}[^{\gls{sup:R}}_{\gls{sub:y}}]$, $\gls{sym:f}[^{\gls{sup:R}}]$, $\gls{sym:f}[^{\gls{sup:V}}]$, $\gls{sym:f}[^{\gls{sup:V2}}]$, by taking $\gls{sym:N}, \gls{sym:M} = 128$ non-periodic samples from each field in order to compare and investigate the present roughness structures in physical space. The line correlation spectra in the \gls{sym:x} and \gls{sym:y}-direction respectively, $\gls{sym:Rtensor}[_{\gls{sub:x}}] (\gls{sym:rvector}[_{\gls{sub:x}}])$ and $\gls{sym:Rtensor}[_{\gls{sub:y}}] (\gls{sym:rvector}[_{\gls{sub:y}}])$, are defined by and determined numerically as expressed by:
\begin{equation*}
\label{eq:CorrelationSpectra}
\begin{split}
        \mathbf{R}_x (\mathbf{r}_x) = \int^{L_x}_0 \left(\overline{f(x, y)f(x + \mathbf{r}_x, y)}\right)' dx \approx \sum^{N-1}_{i = 0} \left(\frac{\overline{f(x_i, y_j)f(x_i + \mathbf{r}_x, y_j)}}{\overline{f(x_i, y_j)}^2}\right) \Delta x,\\
    \mathbf{R}_y (\mathbf{r}_y) = \int^{L_y}_0 \left(\overline{f(x, y)f(x, y + \mathbf{r}_y)}\right)' dy \approx \sum^{M-1}_{j = 0} \left(\frac{\overline{f(x_i, y_j)f(x_i, y_j + \mathbf{r}_y)}}{\overline{f(x_i, y_j)}^2}\right) \Delta y, 
\end{split}
\end{equation*}
where $\gls{sym:rvector}[_{\gls{sub:x}}]$ and $\gls{sym:rvector}[_{\gls{sub:y}}]$ are the separation vectors, defined as $\gls{sub:i}\Delta\gls{sym:x}$ and $\gls{sub:j}\Delta\gls{sym:y}$ respectively, with $\Delta \gls{sym:x}$ and $\Delta \gls{sym:y}$ for non-periodic samples as described in \autoref{sec:Methodology}.

\begin{figure}[h!]
  \centering
    \includegraphics[width=0.8\textwidth]{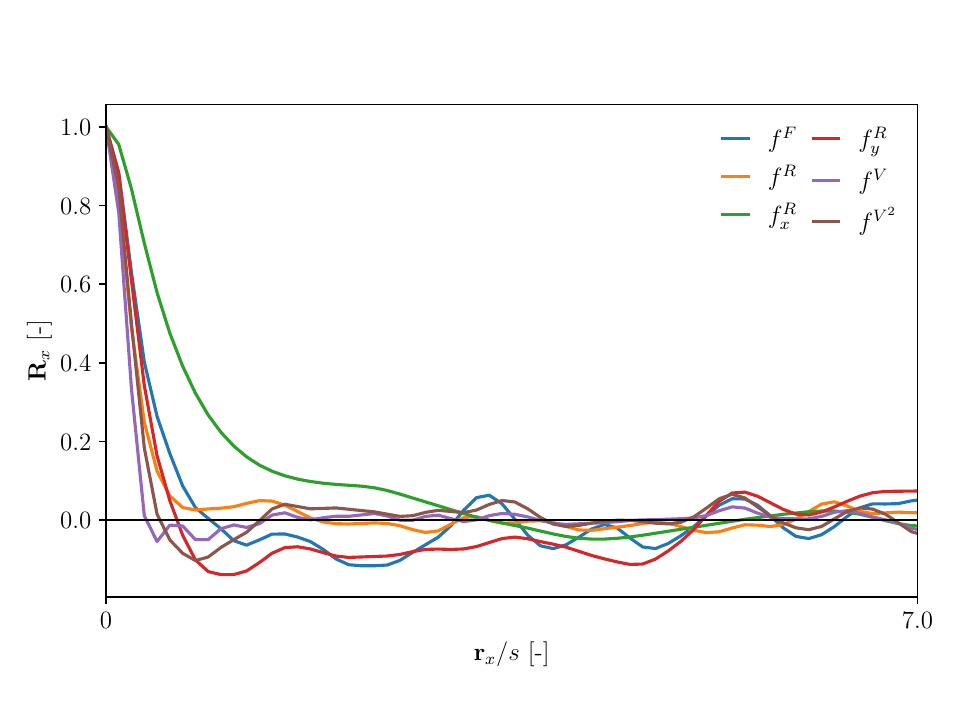}
    \caption{The \gls{sym:x}-two-point correlation spectra $\gls{sym:Rtensor}[_{\gls{sub:x}}]$ comparison of the AM surface \gls{FS} approximation and the five synthetic fields.}
    \label{fig:RxComparison}
\end{figure}

\begin{figure}[h!]
    \centering
     \includegraphics[width=0.8\textwidth]{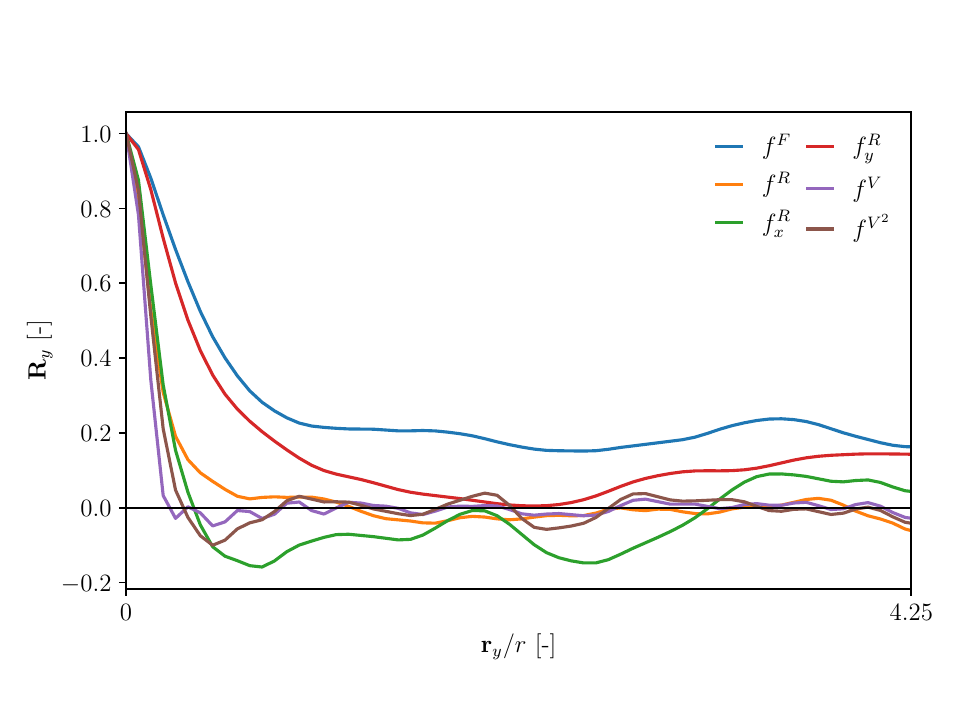}
    \caption{The \gls{sym:y}-two-point correlation spectra $\gls{sym:Rtensor}[_{\gls{sub:y}}]$ comparison of the AM surface \gls{FS} approximation and the five synthetic fields.}
    \label{fig:RyComparison}
\end{figure}

The \gls{sym:x} and  \gls{sym:y}-direction two-point correlation spectra are presented in \autoref{fig:RxComparison} and \ref{fig:RyComparison} respectively. Only the first half of the correlation functions is shown, since all correlation functions in \gls{sym:x} and \gls{sym:y} are symmetric around their midpoint of $\gls{sym:rvector}[_{\gls{sub:x}}]/\gls{sym:s} = 7.0$ and $\gls{sym:rvector}[_{\gls{sub:y}}]/\gls{sym:r} = 4.25$ respectively. The \gls{FS} approximation $\gls{sym:f}[^{\gls{sup:F}}]$ represents the correlations of the \gls{AM} roughness structures. The width at half maximum of the correlation function is a measure of the average size of the contained roughness structures in the fields; all fields with the exception of $f_x^R$ have half-maximum widths close to that of $f_x^F$. Beyond the half-maximum width, for $r_x/s, r_y/r > 1$, the $x-$two-point correlation becomes negative and then oscillates between $-0.1$ and $0.1$; the $y-$two-point correlation, on the other hand, maintains a roughly constant value of $0.2$ up to $r_y/r=4.25$. This can be explained by the ridges in \gls{AM} roughness which follow the laser path - in the present case, the laser path is in the $y-$direction, meaning the roughness height will oscillate in the $x-$direction (leading to oscillations in the two-point correlation) and will tend to remain constant in the $y-$direction (positive correlation plateau). 

Out of the five synthetic fields, $\gls{sym:f}[^{\gls{sup:R}}_{\gls{sub:y}}]$ captures this behavior best, with oscillations in $R_x$ and a consistently positive $R_y$ with a similar half-maximum width (albeit lower values near $r_y/r =2$). Additionally, a qualitative comparison of the roughness field contours shows the highest agreement of $\gls{sym:f}[^{\gls{sup:R}}_{\gls{sub:y}}]$ to $f^F$. This indicates that the optimal Rogallo-based roughness profile is the one based on the component of velocity parallel to the laser's path.

%The roughness structures in the Gaussian roughness model do not closely represent $\gls{sym:f}[^{\gls{sup:F}}]$ completely, since their correlation functions are different. However, the initial drop-off of the $\gls{sym:f}[^{\gls{sup:G}}]$ correlation functions being similar compared to some of the synthetic fields does indicate that approximately the same overall roughness scales are present in $\gls{sym:f}[^{\gls{sup:G}}]$.

\section{Conclusions}
\label{sec:Conclusions}

A data-driven model has been developed for the synthetic generation of Additive Manufacturing (\gls{AM}) roughness fields based on images of \gls{AM} roughness electron microscope scans. The method uses data extraction methods, Fourier analysis and Rogallo's method \cite{Rogallo1981}. The model is well suited for the generation of numerical simulation grids with surface roughness. Five synthetically generated roughness fields based on the Rogallo vector field have been tested. Of those five, the one based on the component of the Rogallo vector field parallel to the laser path performs best, capturing anisotropic features such as differing autocorrelation behaviors along the two axes of the surface. These features are not captured by existing synthetic roughness models, such as those using a random distribution of Gaussian basis functions. Additionally, the Rogallo-based method requires a single image, unlike the 20+ image datasets needed for training of ML algorithms for synthetic roughness. Thus, the Rogallo roughness based on the laser-path-parallel velocity component is proposed as a superior alternative to existing synthetic roughness models in cases when the existing experimental measurements of AM roughness are limited. 
%Mathematical models from literature, such as the Gaussian roughness model, are not able to resemble the roughness elements present in \gls{AM} roughness scans even if the contained roughness scales are similar.
%\input{Sections/Recommendations}
\section{Acknowledgement}

This work is supported by the National Energy Technology Laboratory's University Turbine Systems Research Program, DE-FOA-0002397.

%Bibliography
\bibliographystyle{unsrt}  
\bibliography{references}  

\begin{thebibliography}{10}

\bibitem{Frazier2014}
W.~E. Frazier.
\newblock Metal additive manufacturing: A review.
\newblock {\em Journal of Materials Engineering and Performance}, 23(6):1917--1928, 2014.

\bibitem{BouabbouVaudreuil2022}
A.~Bouabbou and S.~Vaudreuil.
\newblock Understanding laser-metal interaction in selective laser melting additive manufacturing through numerical modelling and simulation: a review.
\newblock {\em Virtual and Physical Prototyping}, 17(3):543--562, 2022.

\bibitem{MowerLong2016}
T.~M. Mower and M.~J. Long.
\newblock Mechanical behavior of additive manufactured, powder-bed laser-fused materials.
\newblock {\em Materials Science and Engineering: A}, 651:198--213, 2016.

\bibitem{AboulkhairEtAl2019}
N.~T. Aboulkhair, M.~Simonelli, L.~Parry, I.~Ashcroft, C.~Tuck, and R.~Hague.
\newblock 3d printing of aluminium alloys: Additive manufacturing of aluminium alloys using selective laser melting.
\newblock {\em Progress in Materials Science}, 106:100578, 2019.

\bibitem{AdeleKniepkamp2015}
E.~Abele and M.~Kniepkamp.
\newblock Analysis and optimisation of vertical surface roughness in micro selective laser melting.
\newblock {\em Surface Topography: Metrology and Properties}, 3(3):034007, 2015.

\bibitem{PatelEtAl2020}
S.~Patel, A.~Rogalsky, and M.~Vlasea.
\newblock Towards understanding side-skin surface characteristics in laser powder bed fusion.
\newblock {\em Journal of Materials Research}, 35(15):2055--2064, 2020.

\bibitem{FoxEtAl2016}
J.~C. Fox, S.~P. Moylan, and B.~M. Lane.
\newblock Effect of process parameters on the surface roughness of overhanging structures in laser powder bed fusion additive manufacturing.
\newblock {\em Procedia CIRP}, 45:131--134, 2016.

\bibitem{KasperovichEtAl2021}
G.~Kasperovich, R.~Becker, K.~Artzt, P.~Barriobero-Vila, G.~Requena, and J.~Haubrich.
\newblock The effect of build direction and geometric optimization in laser powder bed fusion of inconel 718 structures with internal channels.
\newblock {\em Materials \& Design}, 207:109858, 2021.

\bibitem{FaveroEtAl2022}
G.~Favero, G.~Berti, M.~Bonesso, D.~Morrone, S.~Oriolo, P.~Rebesan, R.~Dima, P.~Gregori, A.~Pepato, A.~Scanavini, and S.~Mancin.
\newblock Experimental and numerical analyses of fluid flow inside additively manufactured and smoothed cooling channels.
\newblock {\em International Communications in Heat and Mass Transfer}, 135:106128, 2022.

\bibitem{Chen1972}
K.~C. Chen.
\newblock Compressible turbulent boundary-layer heat transfer to rough surfaces in pressure gradient.
\newblock {\em AIAA Journal}, 10(5):623--629, 1972.

\bibitem{AntoniaLuxton1971}
R.~A. Antonia and R.~E. Luxton.
\newblock The response of a turbulent boundary layer to a step change in surface roughness part 1. smooth-to-rough.
\newblock {\em Journal of Fluid Mechanics}, 48:721 -- 761, 1971.

\bibitem{AntoniaLuxton1972}
R.~A. Antonia and R.~E. Luxton.
\newblock The response of a turbulent boundary layer to a step change in surface roughness. part 2. rough-to-smooth.
\newblock {\em Journal of Fluid Mechanics}, 53:737 -- 757, 1972.

\bibitem{BrezginEtAl2017}
D.~V. Brezgin, K.~E. Aronson, F.~Mazzelli, and A.~Milazzo.
\newblock The surface roughness effect on the performance of supersonic ejectors.
\newblock {\em Thermophysics and Aeromechanics}, 24(4):553--561, 2017.

\bibitem{GramespacherEtAl2021}
C.~Gramespacher, H.~Albiez, M.~Stripf, and H.~J. Bauer.
\newblock The influence of element thermal conductivity, shape, and density on heat transfer in a rough wall turbulent boundary layer with strong pressure gradients.
\newblock {\em Journal of Turbomachinery}, 143(8):9, 2021.

\bibitem{KadivarEtAl2022}
M.~Kadivar, D.~Tormey, and G.~McGranaghan.
\newblock Cfd of roughness effects on laminar heat transfer applied to additive manufactured minichannels.
\newblock {\em Heat and Mass Transfer}, 2022.

\bibitem{McClainEtAl2021}
S.~T. McClain, D.~R. Hanson, E.~Cinnamon, J.~C. Snyder, R.~F. Kunz, and K.~A. Thole.
\newblock Flow in a simulated turbine blade cooling channel with spatially varying roughness caused by additive manufacturing orientation.
\newblock {\em Journal of Turbomachinery-Transactions of the Asme}, 143(7):12, 2021.

\bibitem{FaveroEtAl2021}
G.~Favero, M.~Bonesso, P.~Rebesan, R.~Dima, A.~Pepato, and S.~Mancin.
\newblock Additive manufacturing for thermal management applications: from experimental results to numerical modeling.
\newblock {\em International Journal of Thermofluids}, 10:100091, 2021.

\bibitem{ZhangEtAl2010}
C.~Zhang, Y.~Chen, and M.~Shi.
\newblock Effects of roughness elements on laminar flow and heat transfer in microchannels.
\newblock {\em Chemical Engineering and Processing: Process Intensification}, 49(11):1188--1192, 2010.

\bibitem{NoorianEtAl2014}
H.~Noorian, D.~Toghraie, and A.~R. Azimian.
\newblock The effects of surface roughness geometry of flow undergoing poiseuille flow by molecular dynamics simulation.
\newblock {\em Heat and Mass Transfer}, 50(1):95--104, 2014.

\bibitem{ChenEtAl2010}
Y.~Chen, P.~Fu, C.~Zhang, and M.~Shi.
\newblock Numerical simulation of laminar heat transfer in microchannels with rough surfaces characterized by fractal cantor structures.
\newblock {\em International Journal of Heat and Fluid Flow}, 31(4):622--629, 2010.

\bibitem{DharaiyaKandlikar2013}
V.~V. Dharaiya and S.~G. Kandlikar.
\newblock A numerical study on the effects of 2d structured sinusoidal elements on fluid flow and heat transfer at microscale.
\newblock {\em International Journal of Heat and Mass Transfer}, 57(1):190--201, 2013.

\bibitem{GroceEtAl2007}
G.~Croce, P.~D’agaro, and C.~Nonino.
\newblock Three-dimensional roughness effect on microchannel heat transfer and pressure drop.
\newblock {\em International Journal of Heat and Mass Transfer}, 50(25):5249--5259, 2007.

\bibitem{HuEtAl2003}
Y.~Hu, C.~Werner, and D.~Li.
\newblock Influence of three-dimensional roughness on pressure-driven flow through microchannels.
\newblock {\em Journal of Fluids Engineering, Transactions of the ASME}, 125(5):871--879, 2003.

\bibitem{RawoolEtAl2006}
A.~S. Rawool, Sushanta~K. Mitra, and S.~G. Kandlikar.
\newblock Numerical simulation of flow through microchannels with designed roughness.
\newblock {\em Microfluidics and Nanofluidics}, 2(3):215--221, 2006.

\bibitem{HeckPapavassiliou2013}
M.~L. Heck and D.~V. Papavassiliou.
\newblock Effect of hydrophobicity-inducing roughness on micro-flows.
\newblock {\em Chemical Engineering Communications}, 200(7):919--934, 2013.

\bibitem{Kapsis2019}
M.~Kapsis.
\newblock {\em Multi-Scale CFD Modelling towards Machined Roughness}.
\newblock Doctoral thesis, 2019.

\bibitem{KapsisEtAl2019}
M.~Kapsis, L.~He, Y.~S. Li, O.~Valero, R.~Wells, S.~Krishnababu, G.~Gupta, J.~Kapat, and M.~Schaenzer.
\newblock Multi-scale parallelised cfd modelling towards resolving manufacturable roughness.
\newblock In {\em Proceedings of the ASME Turbo Expo}, volume 2C-2019.

\bibitem{ChenEtAl2009}
Y.~Chen, C.~Zhang, M.~Shi, and G.~P. Peterson.
\newblock Role of surface roughness characterized by fractal geometry on laminar flow in microchannels.
\newblock {\em Physical Review E}, 80(2):026301, 2009.

\bibitem{GuoEtAl2015}
L.~Guo, H.~Xu, and L.~Gong.
\newblock Influence of wall roughness models on fluid flow and heat transfer in microchannels.
\newblock {\em Applied Thermal Engineering}, 84:399--408, 2015.

\bibitem{ZhangEtAl2012}
W.~Zhang, G.~Meng, X.~Wei, and Z.~Peng.
\newblock Slip flow and heat transfer in microbearings with fractal surface topographies.
\newblock {\em International Journal of Heat and Mass Transfer}, 55(23):7223--7233, 2012.

\bibitem{XiongChung2010}
R.~Xiong and Jacob~N. Chung.
\newblock Investigation of laminar flow in microtubes with random rough surfaces.
\newblock {\em Microfluidics and Nanofluidics}, 8(1):11--20, 2010.

\bibitem{SenEtAl2015}
O.~Sen, S.~Davis, G.~B. Jacobs, and H.~S. Udaykumar.
\newblock Evaluation of convergence behavior of metamodeling techniques for bridging scales in multi-scale multimaterial simulation.
\newblock {\em Journal of Computational Physics}, 294:585--604, 2015.

\bibitem{KhorasaniEtAl2018}
A.~M. Khorasani, I.~Gibson, M.~Goldberg, and G.~Littlefair.
\newblock A comprehensive study on surface quality in 5-axis milling of slm ti-6al-4v spherical components.
\newblock {\em International Journal of Advanced Manufacturing Technology}, 94(9-12):3765--3784, 2018.

\bibitem{FotovvatiEtAl2022}
B.~Fotovvati, S.~Rauniyar, J.~A. Arnold, and K.~Chou.
\newblock Experimental, computational, and data-driven study of the effects of selective laser melting (slm) process parameters on single-layer surface characteristics.
\newblock {\em International Journal of Advanced Manufacturing Technology}, 123(1-2):119--144, 2022.

\bibitem{LaFé-PerdomoEtAl2022}
I.~La~Fé-Perdomo, J.~A. Ramos-Grez, I.~Jeria, C.~Guerra, and G.~O. Barrionuevo.
\newblock Comparative analysis and experimental validation of statistical and machine learning-based regressors for modeling the surface roughness and mechanical properties of 316l stainless steel specimens produced by selective laser melting.
\newblock {\em Journal of Manufacturing Processes}, 80:666--682, 2022.

\bibitem{Rogallo1981}
R.~S. Rogallo.
\newblock Numerical experiments in homogeneous turbulence.
\newblock Nasa technical memorandum 81315, National Aeronautics and Space Administration, 1981.

\bibitem{NieuwstadtEtAl2016}
F.~T.~M. Nieuwstadt, B.~J. Boersma, and J.~Westerweel.
\newblock {\em Turbulence: Introduction to Theory and Applications of Turbulent Flows}.
\newblock Springer, 2016.

\bibitem{HanburyEtAl2002}
A.~Hanbury and J.~Serra.
\newblock A 3d-polar coordinate colour representation suitable for image analysis.
\newblock {\em Computer Vision and Image Understanding - CVIU}, 2002.

\bibitem{AltlandEtAl2022}
S.~Altland, X.~Zhu, S.~McClain, R.~Kunz, and X.~Yang.
\newblock Flow in additively manufactured super-rough channels.
\newblock {\em Flow}, 2, 2022.

\end{thebibliography}

%Appendix
%\newpage
%\input{Sections/Appendix}

\end{document}